\documentclass[sigconf]{acmart}
\AtBeginDocument{%
  }

\usepackage{todonotes}
\usepackage{graphicx}
\usepackage{caption}

\usepackage[]{algorithm}
\usepackage{algpseudocode}
\algnewcommand\And{\textbf{and}}
\usepackage{amsfonts}
\usepackage{rotating}
\usepackage{xcolor}
\usepackage{multirow}
\usepackage{color}
\usepackage{array}
\usepackage{booktabs}
\usepackage{tabularx}
\usepackage{multirow}
\usepackage{placeins}
\usepackage{longtable}
\usepackage{supertabular,enumitem}
\usepackage{hyperref}
\usepackage{academicons}
\definecolor{orcidlogocol}{HTML}{A6CE39}
\usepackage{xspace}
\usepackage{todonotes}
\usepackage{booktabs}
\usepackage{amsmath}
\usetikzlibrary{arrows.meta}
\usepackage{eurosym}
\usepackage{diagbox}
\usepackage{subcaption}

\definecolor{inteins}{RGB}{128,179,255}

\newcommand{\MNIST}{\texttt{MNIST}\xspace}
\newcommand{\GAUS}{\texttt{gaussian}\xspace}
\newcommand{\BLOB}{\texttt{blob}\xspace}
\newcommand{\RING}{\texttt{gaussian}\xspace}

\newcommand{\method}{Co-EA}

\begin{document}

%%
%% The "title" command has an optional parameter,
%% allowing the author to define a "short title" to be used in page headers.
\title{Multi-population GAN Training: Analyzing Co-Evolutionary Algorithms}

%%
%% The "author" command and its associated commands are used to define
%% the authors and their affiliations.
%% Of note is the shared affiliation of the first two authors, and the
%% "authornote" and "authornotemark" commands
%% used to denote shared contribution to the research.

\author{Walter P. Casas}
\affiliation{%
  \institution{ITIS UMA, University of Malaga}
  %\city{Hekla}
  \country{Spain}}
\email{walterpcasas@gmail.com}

\author{Jamal Toutouh}
%\authornote{All authors contributed equally to this research.}
\orcid{0000-0001-6923-8445}
\affiliation{%
 \institution{ITIS UMA, University of Malaga}
  \country{Spain}
}
\email{jamal@uma.es}

%\new{This behavior also reflects a trade-off between robustness and precision: $f_1$ favors broader confusion, while $f_2$ encourages more localized misclassification.}
%\new{The smoother curves in MNIST and the noisier patterns in CIFAR-10 also suggest a more rugged fitness landscape in the latter, likely due to higher visual variability.}

% \author{Anon Anon}
% %\authornote{All authors contributed equally to this research.}
% \affiliation{%
%  \institution{Anon Institution}
%   \country{Anon}
% }
% \email{anon@anon.edu}

%%
%% By default, the full list of authors will be used in the page
%% headers. Often, this list is too long, and will overlap
%% other information printed in the page headers. This command allows
%% the author to define a more concise list
%% of authors' names for this purpose.
% \renewcommand{\shortauthors}{Anonymous et al.}
\renewcommand{\shortauthors}{W. P. Casas and J. Toutouh}

%%
%% The abstract is a short summary of the work to be presented in the
%% article.
\begin{abstract}
Generative adversarial networks (GANs) are powerful generative models but remain challenging to train due to pathologies such as mode collapse and instability. Recent research has explored coevolutionary approaches, in which populations of generators and discriminators are evolved, as a promising solution. This paper presents an empirical analysis of different coevolutionary GAN training strategies, focusing on the impact of selection and replacement mechanisms. We compare $(\mu,\lambda)$, $(\mu{+}\lambda)$ with elitism, and $(\mu{+}\lambda)$ with tournament selection coevolutionary schemes, along with a non-evolutionary population based multi-generator multi-discriminator GAN baseline, across both synthetic low-dimensional datasets (\BLOB and \RING mixtures) and an image-based benchmark (MNIST). Results show that full generational replacement, i.e., $( \mu, \lambda )$, consistently outperforms in terms of both sample quality and diversity, particularly when combined with larger offspring sizes. In contrast, elitist approaches tend to converge prematurely and suffer from reduced diversity. These findings highlight the importance of balancing exploration and exploitation dynamics in coevolutionary GAN training and provide guidance for designing more effective population-based generative models.
\end{abstract}

%%
%% The code below is generated by the tool at http://dl.acm.org/ccs.cfm.
%% Please copy and paste the code instead of the example below.
%%


%%
%% Keywords. The author(s) should pick words that accurately describe
%% the work being presented. Separate the keywords with commas.
%% A "teaser" image appears between the author and affiliation
%% information and the body of the document, and typically spans the
%% page.
% \begin{teaserfigure}
%   \includegraphics[width=\textwidth]{sampleteaser}
%   \caption{Seattle Mariners at Spring Training, 2010.}
%   \Description{Enjoying the baseball game from the third-base
%   seats. Ichiro Suzuki preparing to bat.}
%   \label{fig:teaser}
% \end{teaserfigure}

%\received{20 February 2007}
%\received[revised]{12 March 2009}
%\received[accepted]{5 June 2009}

%%
%% This command processes the author and affiliation and title
%% information and builds the first part of the formatted document.
\maketitle

%% Son máximo 8 páginas para GECCO 2025
%% research papers (up to 8 pages)
%% position papers (up to 4 pages)

\section{Introduction}
\label{sec:introduction}
Generative Adversarial Networks (GANs) have emerged as powerful generative models, demonstrating success in image synthesis, data augmentation, and unsupervised representation learning~\cite{ku2023textcontrolgan}. A GAN consists of two artificial neural networks (ANN), a generator ($g$) and a discriminator ($d$), that are trained simultaneously in a competitive setting. The generator aims to produce samples that resemble real training data, while the discriminator attempts to distinguish between real and synthesized samples~\cite{goodfellow2020generative}.

The training process is formulated as a min-max optimization problem~\cite{krawiec2016solving}. The loss functions are designed so that the generator is rewarded for successfully fooling the discriminator, while the discriminator is rewarded for correctly distinguishing between real and fake data samples.

% The training process is formulated as a minimax optimization problem. The loss functions are designed so that the generator is rewarded for successfully fooling the discriminator, while the discriminator is rewarded for correctly distinguishing between real and fake data samples. 

Under ideal conditions, GAN training converges to an equilibrium where the generator produces samples that are indistinguishable from real data. However, in practice, achieving stable convergence remains a significant challenge. Vanishing gradient~\cite{arjovsky2017wasserstein}, mode collapse~\cite{arora2018gans} or discriminator collapse~\cite{li2018limitations} are some of the GAN training pathologies that can arise during the training.  

Several approaches have been proposed to overcome GAN pa\-thol\-o\-gies. Among others, researchers have proposed multi-generator and multi-discriminator architectures, evolutionary-based GAN training, and co-evolutionary approaches. Multi-generator GANs introduce multiple competing generators to enhance sample diversity and mitigate mode collapse, while multi-discriminator frameworks leverage multiple critics to provide more refined and distributed adversarial feedback. Evolutionary algorithms (EAs) have been applied to GAN training to optimize hyperparameters, loss functions, and network architectures through mechanisms such as selection, mutation, and recombination. Recently, co-evolutionary algorithms (Co-EAs) have emerged as a promising paradigm, applying population-based co-evolution to GAN training, where both populations of generators and of discriminators co-adapt over successive iterations. These methods have demonstrated improvements in stability, diversity, and training efficiency.

This work aims to analyze how different selection, replacement, and population dynamics influence Co-EA GAN training stability, diversity, and convergence.
Thus, the main contribution of this work is a systematic evaluation of different coevolutionary strategies for GAN training, focusing on how population dynamics, selection mechanisms, and replacement strategies impact both quality and diversity.
To guide this analysis, we consider the following research questions:
\textbf{RQ1:} How do different coevolutionary strategies—particularly selection and replacement mechanisms—impact the quality and diversity of GAN-generated samples?
\textbf{RQ2:} To what extent does the balance between exploration and exploitation, as controlled by population and offspring sizes, affect the performance and stability of GAN training?
\textbf{RQ3:} Can a simple population based multi-generator multi-discriminator (MG-MD) GAN without selection or replacement be competitive with coevolutionary approaches?

The remainder of the paper is organized as follows. Section~\ref{sec:related-work} introduces the background and related work. 
The evaluated Co-EA GAN training approaches are described in Section~\ref{sec:methods}.
Section~\ref{sec:experimental-settings} describes the experimental methodology and Section~\ref{sec:results} presents the experimental results and discussions. Section~\ref{sec:conclusions} discusses the main conclusions and future work.

% \section{Background and Related Work}
% \subsection{Background}
% \label{sec:background}

% Adversarial Evolutionry Learning (AEL)combines evolutionary strategies with adversarial learning to promote diversity

% Evolutionary Learning selects and varies solutions over time based on their fitness function. The goal is to maximize the fitness function $ \phi : \mathbb{R} \to \mathbb{R} $. A population of candidate solutions at time $t$ is denoted by $P_t$, while the variation operation is $v: \mathbb{R} \to \mathbb{R}$ and the evaluation is $e: \mathbb{R} \to \mathbb{R}$. After applying these operations, new solutions are evaluated and selected according to $ \phi $. Formally, the update can be expressed as $P_t = \phi(e(v(P_{t-1})) \mid \theta)$. The objective is $ \arg\max_{x \in P}\,\phi(e(v(P))\mid\theta)$. Over multiple iterations, this process leads to progressively better solutions.

% Adversarial Learning (AL) involves two explicitly competing adversaries, such as player and opponent, or attacker and defender, represented by $r, b \in \mathbb{R}$. Their engagement is given by the function $f : \mathbb{R} \times \mathbb{R} \to \mathbb{R}$, producing an outcome $y = f(r, b \mid \theta_r, \theta_b)$, where $\theta_r$ and $\theta_b$ are parameters learned by $h : \mathbb{R} \times \mathbb{R} \to \mathbb{R} \times \mathbb{R}$. Often, $\theta_r = \theta_b$. The objective is a minimax game: $\min_{r}\,\max_{b}\,f(r, b \mid \theta)$. One well-known instance of AL is the Generative Adversarial Network (GAN). 

% \subsection{Related Works}
\section{Related Works}
\label{sec:related-work}

Generative Adversarial Networks (GANs) have been used across a wide variety of applications, from image and video synthesis to data enhancement~\cite{ali2024generative}. Despite their remarkable success, training GANs remains complex due to the intricate interaction between the generator and the discriminator, which can lead to undesirable behaviors. This complex interaction frequently results in training pathologies like mode collapse and instability, making these models particularly difficult to train.

Several methods integrate Evolutionary Computation (EC) for GANs to mitigate their training pathologies. One representative example is Evolutionary GAN (EGAN)~\cite{wang2019evolutionary} which employs multiple loss functions, each assigned to a different generator, and selects the best performing generator, introducing diversity. 

Another notable approach is Multi-objective Evolutionary GAN (MO-EGAN)~\cite{baioletti2020multi}, which proposes a framework that simultaneously optimizes conflicting criteria such as fidelity and diversity. In MO-EGAN, a multi-objective evolutionary algorithm is used to evolve the generator, resulting in a more balanced strategy that addresses inherent challenges and enhances both the robustness and quality of the generated outputs.

An additional contribution is Differential-evolution-based GAN (DE-GAN) \cite{zheng2019differential}. This approach applies differential evolution to explore the parameter space in GAN training, specifically targeting edge detection. DE-GAN addresses common training pathologies optimizing the GAN’s parameters, demonstrating the potential of evolutionary strategies to enhance performance in specialized tasks.

Other approaches have shown promise in addressing inherent adversarial challenges. Adversarial Evolutionary Learning (AEL), implemented as competitive CoEA~\cite{zhao2021coea}, has proven effective for search, optimization, design and modeling \cite{Antonio2018, Popovici2012, rosin1997new, sims1994evolving}. Its applications include board games~\cite{pollack1998co}, video game playing \cite{keaveney2011evolving, sipper2011evolved}, social science games~\cite{axelrod1981evolution}, software engineering tasks as coevolving programs and unit tests \cite{arcuri2007coevolving, arcuri2014co, wilkerson2010coevolutionary} or differentiating correct from incorrect behavior \cite{barr2014oracle}, and various cybersecurity problems~\cite{hingston2011red, rush2015coevolutionary, service2009increasing}.

As GANs represent a type of adversarial learning, some authors have successfully applied Competitive Co-EA to train GANs.
COEGAN \cite{costa2020neuroevolution, costa2019coegan} takes to a coevolutionary approach to jointly evolve the architectures of generator and discriminator, and t-SNE visualization analysis suggest convergence to high quality model \cite{costa2021demonstrating}. Numerous extensions of this idea include multi objective variants, based on NSGA-II \cite{baioletti2020multi},  quality diversity methods. Differential evolution approaches and cooperative CoEA that evolve both generators and discriminators for multi-objective optimization. Other works replace the generator with an expression-based population.
Few works explore spatial distributed CoEA for GAN training, where adversaries are arranged within a spatial neighborhood or grid. Spatial organization can provide advantages in diversity maintenance, hyperparameter exploration and fine-grained interactions during training~\cite{hemberg2021spatial,toutouh2020parallel,toutouh2019spatial,toutouh2020data,toutouh2021signal,toutouh2023semi}.

Most evolutionary GANs applications focus on computer vision tasks such as hyperspectral image classification~\cite{bai2022immune}, abnormal electrocardiogram classification~\cite{wang2022evolving}, data augmentation for cardiac magnetic resonance image~\cite{fu2021evolutionary}, or COVID-19 infection segmentation~\cite{he2021evolvable, toutouh2020parallel,flores2022coevolutionary}. Beyond vision, evolutionary GAN have been employed for natural language generation~\cite{sun2021composite}, sequential data imputation~\cite{chakraborty2021sequential}, and evolving the GAN architectures of generators and discriminators~\cite{costa2019coevolution, garciarena2018evolved, ying2022eagan}.

\section{GAN Training Methods Evaluated}
\label{sec:methods}

This section introduces the standard GAN and the multi-population GAN training approaches evaluated in this paper: multi-generator multi-discriminator GAN (MG-MD GAN), ($\mu{,}\lambda$) Co-EA GAN, and ($\mu{+}\lambda$) Co-EA GAN.
%   These strategies differ in how they use or omit evolutionary mechanisms to update the generator and discriminator populations.

Standard GAN trains a single pair of artificial neural networks (ANNs) named generator $G$ and discriminator $D$ using adversarial learning. In this framework, during the iterative training process, the generator $G$ attempts to produce realistic samples to deceive the discriminator, while $D$ aims to distinguish between real samples from the training data set and produced samples created by $G$. Equation~\ref{eq:gan-training} shows the mathematical formulation proposed by Goodfellow et al.~\cite{goodfellow2020generative}. Here, $p_{\text{real}}$ denotes the real data distribution and $p_z$ is a prior distribution over the latent space.

\begin{equation}
\min_{G} \max_{D} V(D, G) = \mathbb{E}_{x \sim p_{\text{real}}}[\log D(x)] + \mathbb{E}_{z \sim p_{z}}[\log (1 - D(G(z)))]
\label{eq:gan-training}
\end{equation}

MG-MD GAN incorporates multiple generators and discriminators organized as fixed populations. However, unlike evolutionary methods, there is no selection or replacement during training. During the training process, for each iteration, the generators and discriminators are randomly coupled. Then, each generator-discriminator pair is trained independently using gradient descent according to Equation~\ref{eq:gan-training} for a fixed number $n_t$ of epochs. After a given stop criterion, at the end of the training, the best generator and discriminator are selected based on a predefined quality metric. This setup allows isolating the impact of population diversity from evolutionary dynamics~\cite{toutouh2020analyzing}.

In ($\mu{,}\lambda$) Co-EA GAN, evolutionary dynamics are introduced through a $(\mu,\lambda)$ Co-EA. Two evolving populations are maintained: one for generators and one for discriminators. At each generation, $\lambda$ individuals are selected from each population using a fitness-based selection process, e.g., tournament selection, (Algorithm~\ref{alg:coea}, lines~\ref{alg:select-g}–\ref{alg:select-d}). These selected individuals form the offspring populations $\mathbf{g}^*$ and $\mathbf{d}^*$.

\begin{algorithm}[!ht]
    \footnotesize
    \captionsetup{font=small}
    \caption{\method~training 
            \newline
		\textbf{Input:}
                $T_B$: Computational budget (training epochs), 
                $p_{real}$: Training dataset,
                $B_s$: Batch size, 
                \hbox{$\theta_{g}$, $\theta_{d}$: Initial generator and discriminator parameters},
                $\mu$: Population size,
                $\lambda$: Offspring size,
                $\tau$: Selection parameters,
                $n_e$: Number of evaluation batches,
                $n_t$: Number of training epochs per generator-discriminator couple
            \newline		
		\textbf{Return:}
    		~$g,d$: Trained generator and discriminator
            }
    \label{alg:coea}
    
    \begin{algorithmic}[1]
        \State $\mathbf{g} \gets$ initializePopulation($\theta_{g}$) \Comment{Initialize population $\mathbf{g}$}\label{alg:initialization-g}
        \State $\mathbf{d} \gets$ initializePopulation($\theta_{d}$) \Comment{Initialize population $\mathbf{d}$}\label{alg:initialization-d}
        \State $\iota \gets \left\lfloor\frac{T_B}{n_t  \lambda} \right\rfloor$ \Comment{Compute the number of generations to perform}
	\State $\mathcal{L}_{g,d} \gets$ evaluate($\mathbf{g}, \mathbf{d}, n_e$) \Comment{Evaluate populations} \label{alg:fit-eval-1}
	\For{$i=1$ \textbf{to} $\iota$} \Comment{Loop over generations}  
            \State $\mathbf{g}^* \gets$ select($\lambda, \tau, \mathcal{L}_{g,d}$) \Comment{Selection to get offspring population $\mathbf{g}^*$} \label{alg:select-g}
            \State $\mathbf{d}^* \gets$ select($\lambda, \tau, \mathcal{L}_{g,d}$) \Comment{Selection to get offspring population $\mathbf{d}^*$} \label{alg:select-d} 
            \State $\mathbf{d}' \gets \mathbf{d}^*$ \Comment{Initialize the set of disc. to couple with gen.}
		\For{$g \in \mathbf{g}^*$} \Comment{Loop over $g$ in the offspring $\mathbf{g}^*$}
                \State $d \gets $ pickOneRandomly($\mathbf{d}'$) \Comment{Select a discriminator for $g$}
                \State $\mathbf{d}' \gets \mathbf{d}' - \{d\}$ \Comment{Remove the selected $d$ from the set}
		      \State $g, d \gets$ train($g, d, n_t, p_{real}, B_s, \theta_g, \theta_d$) \Comment{Train} \label{alg:method-ssl-training}
		\EndFor
            \State $\mathbf{g} \gets \mathbf{g} \cup \mathbf{g}^*$ \Comment{Add offspring to population} \label{alg:add-offspring-g}
            \State $\mathbf{d} \gets \mathbf{d} \cup \mathbf{d}^*$ \Comment{Add offspring to population} \label{alg:add-offspring-d}
		\State $\mathcal{L}_{g,d} \gets$ evaluate($\mathbf{g}, \mathbf{d}, n_e$) \Comment{Evaluate populations} \label{alg:fit-eval-2}
            \State $\mathbf{g} \gets$ updatePopulation($\mathbf{g}$) \Comment{Apply replacement} \label{alg:update-pops-g}
		\State $\mathbf{d} \gets$ updatePopulation($\mathbf{d}$) \Comment{Apply replacement} \label{alg:update-pops-d}
        \EndFor
        \State $g \gets$ selectBestGenerator($\mathbf{g}$)  \label{alg:select-best-g}
        \State $ d \gets$ selectBestDiscriminator($\mathbf{d}$)  \label{alg:select-best-d}
	\State \Return $g$, $d$ \Comment{Return $g$, $d$ trained SSL-GAN}\label{alg:return}
    \end{algorithmic}
\end{algorithm}

Each offspring generator is randomly paired with a discriminator from the offspring discriminator pool (lines~\ref{alg:select-g}–\ref{alg:method-ssl-training}) and trained for $n_t$ epochs using gradient descent. After training, the offspring are added to the current populations, which in the case of $(\mu,\lambda)$ is empty, (lines~\ref{alg:add-offspring-g}–\ref{alg:add-offspring-d}), and fitness is re-evaluated (line~\ref{alg:fit-eval-2}).

In the $(\mu,\lambda)$ strategy, the final population update step (lines~\ref{alg:update-pops-g}–\ref{alg:update-pops-d}) discards the parent population and retains only the $\mu$ best individuals from the $\lambda$ offspring, enforcing full generational replacement and promoting exploration.

The $(\mu{+}\lambda)$ Co-EA GAN introduces elitism by modifying the replacement strategy. Here, the population update selects the top $\mu$ individuals from the union of parents and offspring. This approach preserves high-performing individuals from previous generations, which encourages exploitation of learned solutions while maintaining evolutionary innovation.

This variant shares the same initialization, selection, training, and evaluation phases as the $(\mu,\lambda)$ Co-EA GAN (Algorithm~\ref{alg:coea}), but differs in how the replacement is performed in lines~\ref{alg:update-pops-g}–\ref{alg:update-pops-d}. The function \texttt{updatePopulation()} is adapted to select the $\mu$ best individuals from $\mathbf{g} \cup \mathbf{g}^*$ and $\mathbf{d} \cup \mathbf{d}^*$ rather than just the offspring.

\section{Experimental Setup}
\label{sec:experimental-settings}

\begin{figure}
\begin{subfigure}[t]{0.3\linewidth}
    \centering
    \includegraphics[width=\linewidth]{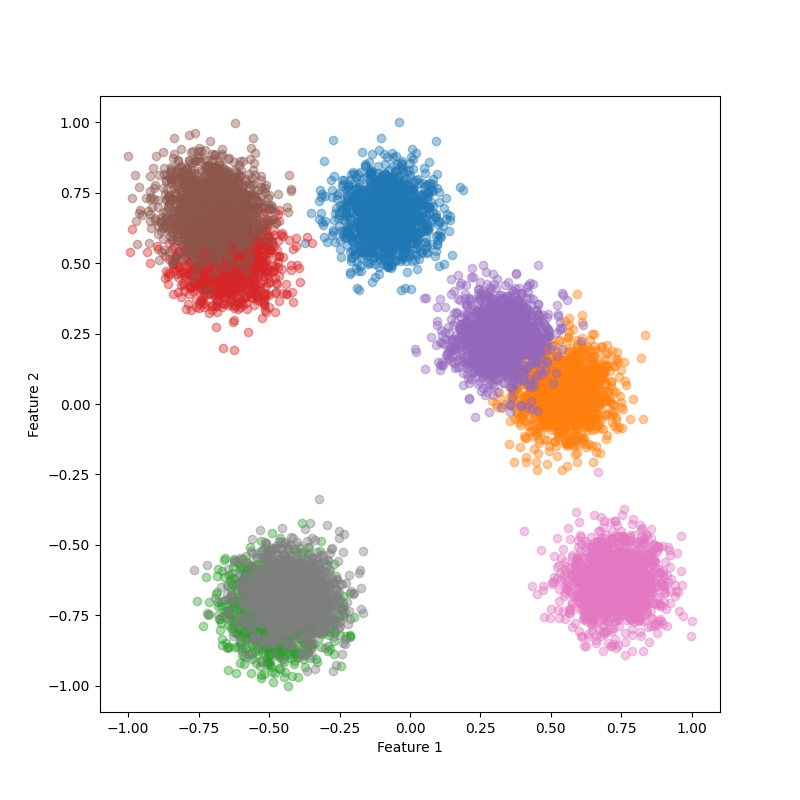}
    \caption{\BLOB-8}
    \label{fig:blobs-dataset}
\end{subfigure}
\begin{subfigure}[t]{0.3\linewidth}
    \centering
    \includegraphics[width=\linewidth]{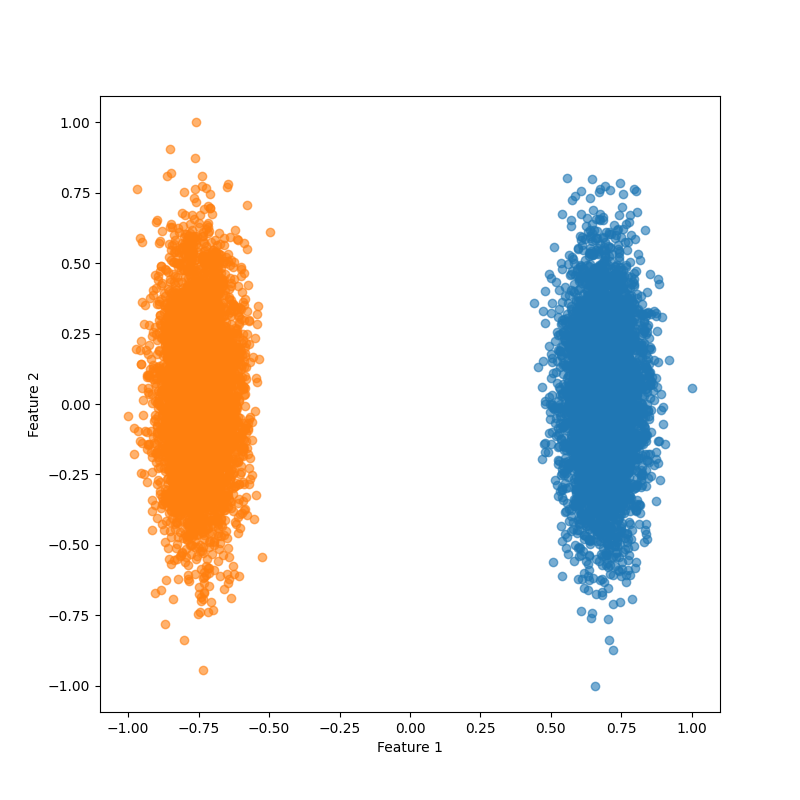}
    \caption{\RING-2}
    \label{fig:ring-dataset-2}
\end{subfigure}
\begin{subfigure}[t]{0.3\linewidth}
    \centering
    \includegraphics[width=\linewidth]{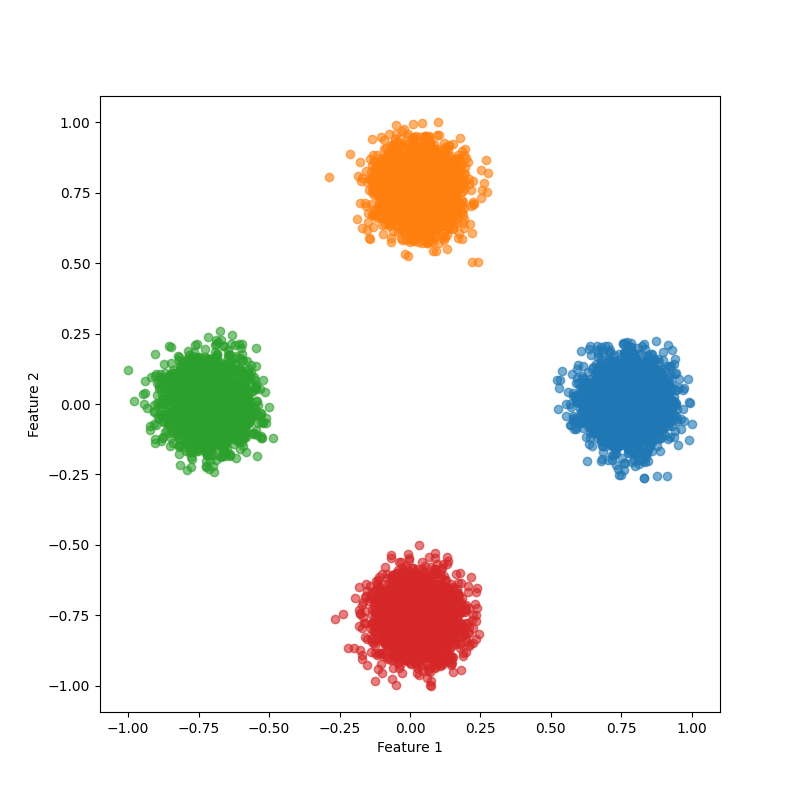}
    \caption{\RING-4 }
    \label{fig:ring-dataset4}
\end{subfigure}

\vspace{0.2cm}

\begin{subfigure}[t]{0.3\linewidth}
    \centering
    \includegraphics[width=\linewidth]{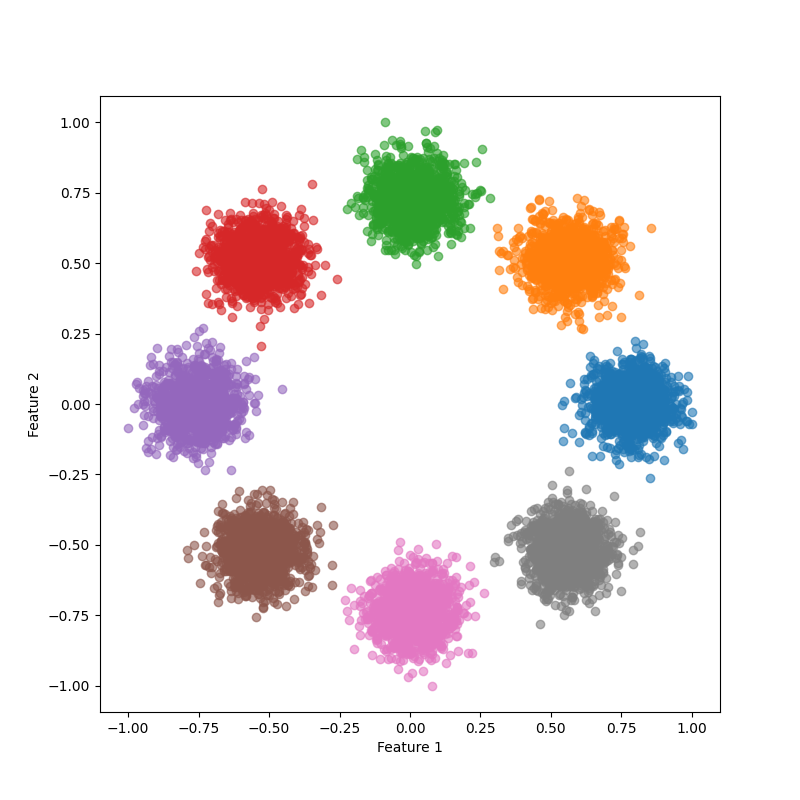}
    \caption{\RING-8 }
    \label{fig::ring-dataset-8}
\end{subfigure}
\begin{subfigure}[t]{0.3\linewidth}
    \centering
    \includegraphics[width=\linewidth]{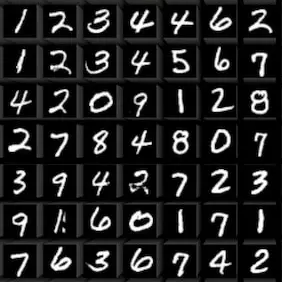}
    \caption{\MNIST }
    \label{fig::mnist-dataset}
\end{subfigure}
    \caption{Representation of the datasets applied in our empirical analysis.}
    \label{fig:datasets}
\end{figure}

The experimental evaluation is performed on three datasets: 
a mixture of eight 2D Gaussian distributions randomly arranged, referred to as \BLOB; synthetic datasets comprising two, four, and ten 2D Gaussian modes arranged uniformly in a circle, referred to as \RING-2, \RING-4, and \RING-8, respectively; and the standard Modified National Institute of Standards and Technology database, \MNIST~\cite{deng2012mnist}. Figure~\ref{fig:datasets} provides visual examples of the used datasets.

Both \BLOB and \RING datasets consist of 2D vectors within the range $[-1, 1]$ and are divided into ten clusters (i.e., Gaussian distributions), forming a total of 10,000 training samples and 1,000 test samples. Although simple in dimensionality, these datasets are widely adopted in generative modeling research due to the challenges they pose—particularly in terms of capturing diversity without collapsing into a subset of the modes.

Although both the \RING and \BLOB datasets are composed of mixtures 2D Gaussians, they present different challenges from a generative modeling point of view. The \RING dataset features a number of well-separated Gaussian modes arranged symmetrically in a circular pattern (see figures~\ref{fig:ring-dataset-2},\ref{fig:ring-dataset4}, and~\ref{fig::ring-dataset-8}). While this structure makes the distribution visually interpretable, it also imposes a stricter requirement on the generator to produce samples across all distinct modes. This characteristic makes \RING particularly susceptible to mode collapse, where the generator may focus on a subset of the modes and ignore the rest.

On the other hand, the \BLOB dataset contains eight Gaussian modes positioned randomly in the feature space, often with some degree of overlap between them (see Figure~\ref{fig:blobs-dataset}). This lack of clear separation reduces the penalty for missing certain regions of the data distribution, making the generation task comparatively easier. Since the modes are not uniformly distributed, even generators that fail to model all components accurately may still produce samples that appear plausible under the data distribution. 

The \MNIST dataset comprises grayscale images of handwritten digits from 0 to 9, each with a resolution of 28$\times$28 pixels. The training partition includes 60,000 samples, while the test set contains 10,000 images. Due to its complexity and high-dimensional nature, \MNIST serves as a standard benchmark for evaluating the generative quality and diversity of image-based models.

In our experiments, the performance of the evaluated methods is evaluated in terms of sample quality and diversity. For \BLOB and \RING datasets, quality is assessed using the Wasserstein Distance ($W_D$)~\cite{vallender1974calculation}, which measures distributional discrepancies in a 2D space. Diversity is quantified through mode coverage~\cite{metz2016unrolled} and cluster-based entropy~\cite{ambrogioni2021automatic}, which respectively capture how many Gaussian modes are represented and how evenly samples are distributed among them, respectively.

In the \MNIST experiments, sample quality is measured using the Fréchet Inception Distance (FID)~\cite{fid}, a standard metric for visual realism based on deep feature statistics. Diversity is evaluated using Total Variation Distance (TVD)~\cite{goodfellow2020generative} between label distributions of real and generated samples, along with mode coverage over the ten digit classes.

The experiments are conducted using two distinct architectures tailored to the datasets under study. 
A simple multilayer perceptron (MLP)-based architecture is used to address \BLOB and \RING data generation and a convolutional neural network (CNN) is used produce \MNIST.

In the case of \BLOB and \RING, the generator is a three-layer MLP that transforms a latent vector into a 2D point. It includes a hidden layer with 128 ReLU units and a final $\tanh$ activation to constrain outputs to $[-1, 1]$. The discriminator is a three-layer MLP with a hidden layer with 128 LeakyReLU units and a sigmoid unit outputs a binary classification (real/fake).

For \MNIST, the generator begins with a fully connected layer mapping the latent vector into a 7$\times$7 feature map, which is then upsampled through two transposed convolutional layers with ReLU activations and batch normalization, producing a final 28$\times$28 grayscale image. The discriminator consists of four convolutional blocks that downsample the input, followed by a sigmoid-based binary \hbox{classifier}.

The experimental design focuses on assessing the impact of key factors in a co-evolutionary algorithm: population sizes, offspring, and selection/replacement strategy. For the $(\mu{+}\lambda)$ variant, we applied elitism and tournament selection, testing population sizes $\mu \in \{3, 5, 7\}$, and offspring sizes $\lambda \in \{1, \lceil \mu/2 \rceil, \mu\}$. The $(\mu{+}\lambda)$ variant that applies elitism is named $(\mu{+}\lambda)_E$ and the variant that uses tournament selection is referred as $(\mu{+}\lambda)_T$. For the $(\mu,\lambda)$ strategy, we also employed tournament selection, with $\mu \in \{3, 5, 7\}$ and $\lambda \in \{\mu, \lceil 1.5\mu \rceil, 2\mu\}$. The choice of $\lambda$ values in each case was made to ensure that the effective selective pressure when generating the next population remained comparable between both methods. The static coevolution baseline (without selection or replacement) was evaluated using fixed population sizes of 3, 5, and 7.

A set of preliminary experiments were performed to confirm the main hyper-parameters used to train the GANs proposed by Sedeño et al.~\cite{sedeno2025}.
Generators and discriminators apply the Adam optimizer with the same learning rate 0.0003. The batch size is 100 samples for \BLOB and \RING and 600 for \MNIST. The total training epochs $T_B$ is set to $250\lambda$ and the number of training epochs $n_t$ when coupling generator-discriminator is 5.

\section{Experimental Results}
\label{sec:results}
In this section, we present the results of the experiments, split by type of datasets, and discusses the research questions. 
Thirty independent runs were performed in all the cases. 

%In each dataset, we first analyze the influence of the number of training epochs during offspring generation, $n_t$, and then the population and the size of the offspring, $\mu$ and $\lambda$, respectively (\textbf{RQ1}). We also show the results obtained by a standard SSL-GAN (\textbf{RQ2}). , except in the analysis of the sensitivity of the methods to the number of labels (\textbf{RQ3}) in \MNIST, where 15 independent runs were performed.

\subsection{Results on \BLOB and \RING}

\begin{figure*}[!ht]
\begin{subfigure}[t]{0.9\linewidth}
    \centering
    \includegraphics[width=0.8\linewidth]{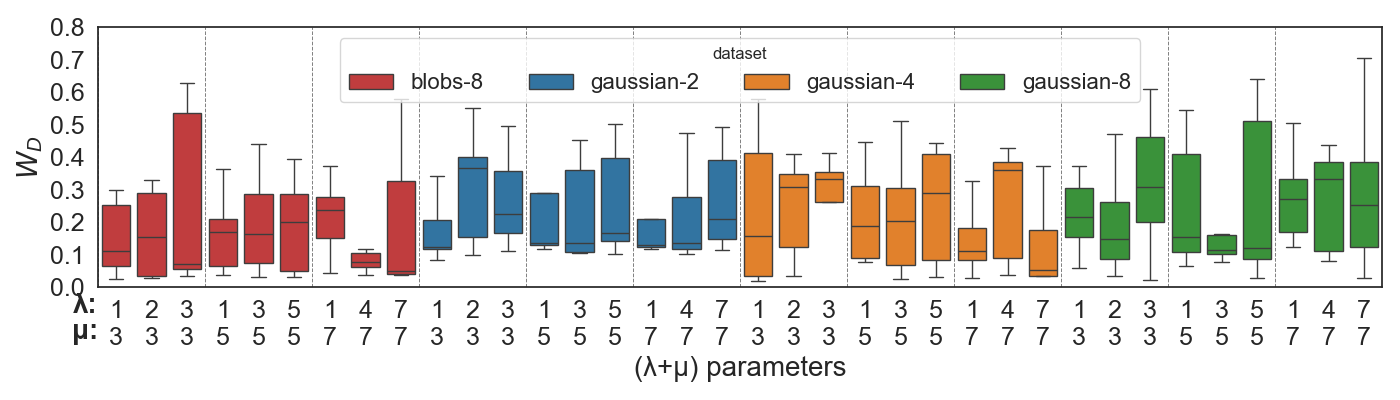}
        \vspace{-0.8em} % Reduce space between image and caption
    \caption{$W_D$ results for $(\mu+\lambda)_{E}$ experiments on the \BLOB and \GAUS datasets.}
    \label{fig:ring-accuracy-population}
\end{subfigure}%

\begin{subfigure}[t]{0.9\linewidth}
   \centering
    \includegraphics[width=0.8\linewidth]{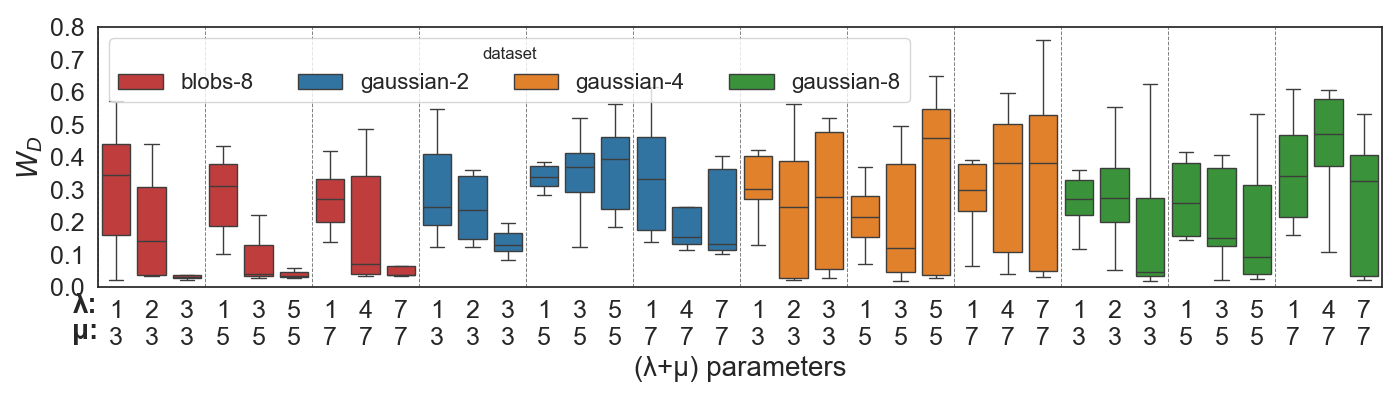}
    \vspace{-0.8em} % Reduce space between image and caption
    \caption{$W_D$ results for $(\mu+\lambda)_{T}$ experiments on the \BLOB and \GAUS datasets.}
    \label{fig:ring-wd-population}
\end{subfigure}

\begin{subfigure}[t]{0.9\linewidth}
   \centering
    \includegraphics[width=0.8\linewidth]{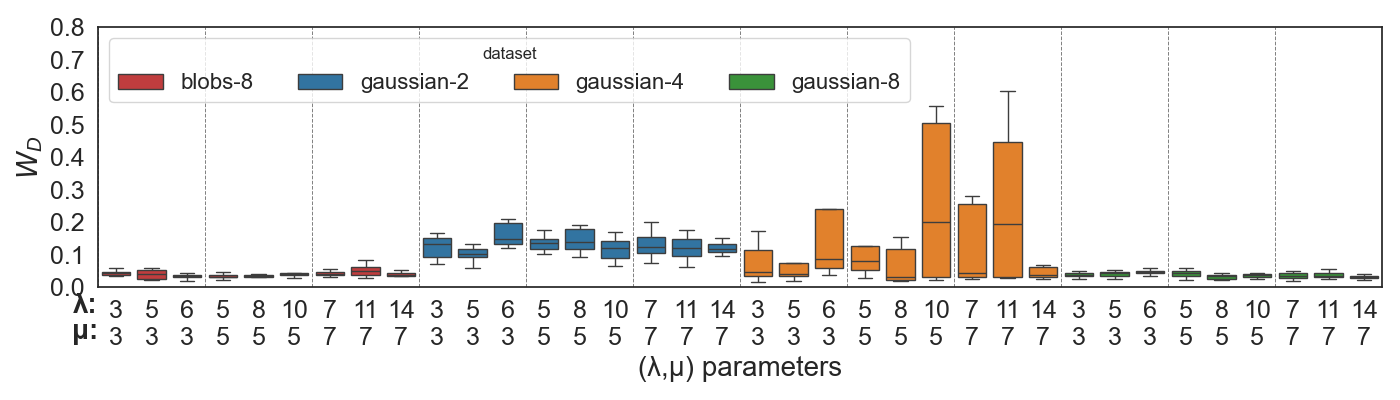}
    \vspace{-0.8em} % Reduce space between image and caption
        \caption{$W_D$ results for $(\mu,\lambda)$ experiments on the \BLOB and \GAUS datasets.}
    \label{fig:ring-wd-population}
\end{subfigure}

\begin{subfigure}[t]{0.9\linewidth}
   \centering
    \includegraphics[width=0.8\linewidth]{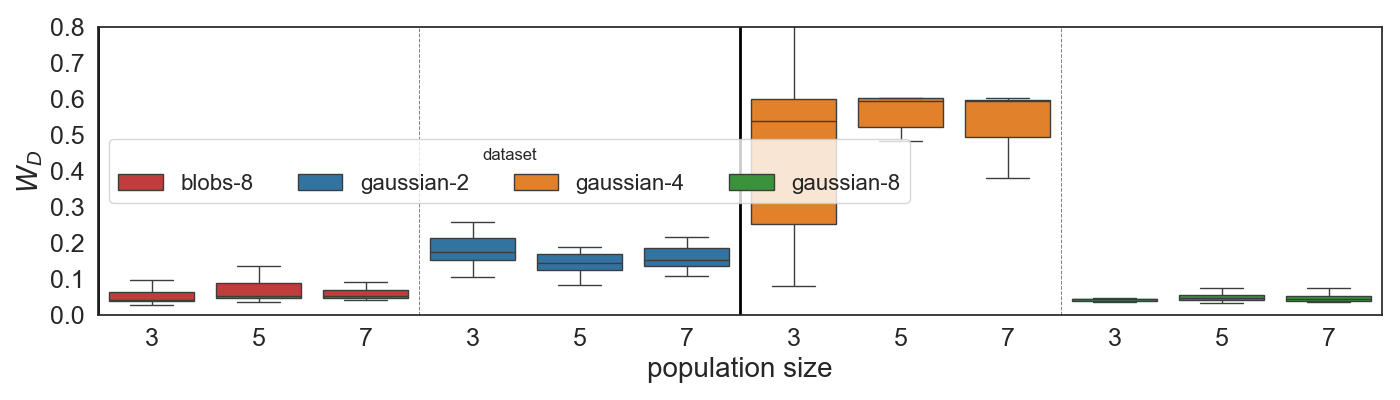}
    \vspace{-0.8em} % Reduce space between image and caption
    \caption{$W_D$ results for MG-MD GAN experiments on the \BLOB and \GAUS datasets.}
    \label{fig:ring-wd-population}
\end{subfigure}
\caption{Influence of $\mu$ and $\lambda$ size in $W_D$ on the \BLOB and \GAUS datasets.}
\label{fig:quality-populations}
\end{figure*}

Figure~\ref{fig:quality-populations} presents the distribution of $W_D$ scores across all evaluated methods considering the different $\mu$ and $\lambda$ on the \RING and \BLOB datasets.

In general, Figure~\ref{fig:quality-populations} shows that among the evaluated datasets, \RING-4 consistently exhibits the highest $W_D$ values, which indicates it is the most challenging setting for all methods. This can be attributed to the structured and symmetric arrangement of modes in the \RING dataset, which imposes strong constraints on the generator to uniformly cover all regions. Any mode collapse or imbalance in sample distribution results in a significant penalty under $W_D$. In contrast, the \BLOB dataset yields the lowest $W_D$ scores across methods, reflecting its relative ease for generative modeling. %Due to the random and partially overlapping mode arrangement in \BLOB, deviations from full mode coverage are less penalized, making the generation task more tolerant to local mode failures.

When comparing the $(\mu{+}\lambda)$ variants, Figure~\ref{fig:quality-populations} shows that $(\mu{+}\lambda)_E$ underperforms $(\mu{+}\lambda)_T$. Elitism reduces exploratory pressure by preserving top individuals across generations, which can lead to early convergence and limited diversity in the population. This lack of diversity appears to hinder the generator's ability to capture all data modes, especially in more structured distributions like \RING. Moreover, in elitist settings, increasing the population or offspring size does not lead to quality improvements, suggesting that the preserved elite dominates the evolutionary dynamics, reducing the effect of new individuals. In contrast, tournament-based selection shows a clear improvement in $W_D$ scores as the offspring size increases. This can be attributed to the higher selective pressure and greater exploration of the search space enabled by larger offspring pools. While increasing the population size also has a positive effect, the results suggest that offspring size plays a more critical role, likely because it directly controls the volume of candidate solutions explored in each generation.

The $(\mu,\lambda)$ variation provides the most competitive $W_D$ results. It also exhibits greater robustness, as reflected by narrower interquartile ranges in Figure~\ref{fig:quality-populations}. While improvements are observed with increasing population and offspring sizes, these gains are generally limited and not always statistically significant according to the Wilcoxon rank-sum test. A key factor behind the effectiveness of $(\mu,\lambda)$ is its stronger capacity to preserve diversity across generations, as offspring completely replace the parent population. This strategy encourages broader exploration of the search space and reduces the risk of premature convergence—a phenomenon previously discussed in the context of evolutionary optimization and coevolutionary GANs~\cite{hemberg2021spatial}. 

The baseline, MG-MD GAN, without selection or replacement also shows competitive results, outperforming $(\mu{+}\lambda)_E$ and approaching the best $(\mu{+}\lambda)_T$ results. These findings highlight the sensitivity of coevolutionary GAN training to the choice of evolutionary strategy and underscore the importance of balancing exploration and exploitation dynamics in population-based training.

To analyze the diversity of the generated samples, we report results using two complementary metrics: cluster-based entropy, which measures how evenly the generator distributes samples across clusters, and mode coverage, which counts how many distinct Gaussian modes are captured by the generated distribution. Tables~\ref{tab:entropy-mu-plus-lambda} and~\ref{tab:entropy-mu-lambda} summarize the cluster-based entropy for $(\mu{+}\lambda)$ and $(\mu,\lambda)$ variants, respectively, and Tables~\ref{tab:modes-mu-plus-lambda} and~\ref{tab:modes-mu-lambda} present the corresponding mode coverage results. MG-MD GAN baseline diversity metrics are also reported for comparison. These tables present median values and interquartile range (iqr) on 30 independent runs. Due to space limitations, we discuss the diversity for the datasets that showed the extreme cases in terms of $W_D$, i.e., \BLOB-8 (best $W_D$ results) and \RING-4 (worst $W_D$ results). 

\begin{table}[h!]
\captionsetup{position=top, skip=2pt}
\renewcommand{\arraystretch}{0.9}  % or 0.8 for more compact

\setlength{\tabcolsep}{3pt} % Default is 6pt
    \centering
    \caption{Cluster-based entropy median (iqr) for $(\mu+\lambda)_E$ and $(\mu+\lambda)_T$ on \BLOB-8 and \RING-4. The lower the better.} \label{tab:entropy-mu-plus-lambda}
    \begin{tabular}{
        c@{\hskip 4pt}
        c@{\hskip 6pt}c
        c@{\hskip 4pt}
        c@{\hskip 10pt}
        c@{\hskip 4pt}
        c
    }
        \toprule
        & & \multicolumn{2}{c}{\BLOB-8} &\phantom{-} & \multicolumn{2}{c}{\RING-4} \\
        \cmidrule{3-4} \cmidrule{6-7}  
        $\mu$ & $\lambda$ 
        & $(\mu+\lambda)_E$ & $(\mu+\lambda)_T$ &\phantom{-}
        & $(\mu+\lambda)_E$ & $(\mu+\lambda)_T$ \\
        
        \midrule
        3 & 1 & 1.513 (1.290) & 1.257 (0.606) &\phantom{-} & 0.693 (1.384) & 0.582 (0.692) \\
        3 & 2 & 1.675 (0.914) & 1.946 (1.105) &\phantom{-} & 0.693 (1.383) & 0.691 (1.384) \\
        3 & 3 & 1.917 (0.401) & 1.992 (0.035) &\phantom{-} & 0.693 (0.693) & 0.692 (1.365) \\
        5 & 1 & 1.751 (0.378) & 1.199 (0.486) &\phantom{-} & 0.693 (0.905) & 0.429 (0.682) \\
        5 & 3 & 1.843 (0.321) & 1.943 (0.662) &\phantom{-} & 0.693 (0.208) & 0.344 (1.378) \\
        5 & 5 & 1.572 (0.849) & 1.956 (0.194) &\phantom{-} & 0.692 (0.693) & 0.346 (1.384) \\
        7 & 1 & 1.630 (0.272) & 1.131 (0.328) &\phantom{-} & 0.692 (0.258) & 0.512 (0.692) \\
        7 & 4 & 1.838 (0.732) & 1.803 (1.296) &\phantom{-} & 0.693 (1.122) & 0.692 (0.693) \\
        7 & 7 & 1.909 (0.168) & 1.917 (0.659) &\phantom{-} & 0.693 (1.385) & 0.000 (0.773) \\
        \bottomrule
    \end{tabular}
\end{table}

\begin{table}[h!]
\captionsetup{position=top, skip=2pt}.
\renewcommand{\arraystretch}{0.9}  % or 0.8 for more compact

    \centering
    \caption{Cluster-based entropy for $(\mu,\lambda)_E$ on \BLOB-8 and \RING-4 in terms of median (iqr). The lower the better.} \label{tab:entropy-mu-lambda}
    \begin{tabular}{cccc}
        \toprule
        $\mu$ & $\lambda$ & \BLOB-8 & \RING-4 \\
        \midrule
        3 & 3 & 2.018 (0.130) & 1.235 (1.384) \\
        3 & 5 & 1.966 (0.115) & 1.383 (0.148) \\
        3 & 6 & 2.012 (0.070) & 1.346 (0.417) \\
        5 & 5 & 2.003 (0.086) & 1.242 (0.378) \\
        5 & 8 & 1.999 (0.075) & 1.375 (0.867) \\
        5 & 10 & 1.973 (0.049) & 1.322 (1.384) \\
        7 & 7 & 1.970 (0.047) & 1.383 (0.324) \\
        7 & 11 & 1.979 (0.073) & 1.378 (1.039) \\
        7 & 14 & 1.981 (0.052) & 1.376 (0.644) \\
        \bottomrule
    \end{tabular}
\end{table}

In terms of cluster-based entropy, the $(\mu, \lambda)$ strategy exhibits the most stable and consistent behavior across both \BLOB-8 and \RING-4 datasets. As shown in Table~\ref{tab:entropy-mu-lambda}, it achieves median entropy values close to 2.0 on \BLOB-8 with very low interquartile ranges (e.g., 1.973 (0.049) for $\mu=5$, $\lambda=10$), indicating a balanced spread of generated samples across all clusters. On the more challenging \RING-4, entropy values are higher and more dispersed, but still consistently outperform the elitist $(\mu{+}\lambda)_E$ approach and match or surpass tournament-based variants.

The $(\mu{+}\lambda)_T$ strategy achieves moderate improvements over its elitist counterpart $(\mu{+}\lambda)_E$ in some configurations, especially for lower values of $\mu$ and $\lambda$. For instance, in \BLOB-8 with $\mu=5$, $\lambda=1$, $(\mu{+}\lambda)_T$ yields an entropy of 1.199 (0.486), outperforming the corresponding elitist version (1.751 (0.378)) (see Table~\ref{tab:entropy-mu-plus-lambda}). However, as population and offspring sizes increase, both $(\mu{+}\lambda)$ variants tend to converge to similar or worse entropy values, with wider dispersion, suggesting that neither approach reliably maintains diversity without a full replacement scheme.

MG-MD GAN achieves reasonably low entropy values on \BLOB-8, median (iqr) results are: 1.950 (0.062), 1.924 (0.123), and 1.920 (0.073) for population sizes 3, 5, and 7, respectively. These results are close to the $(\mu{+}\lambda)_E$ and $(\mu{+}\lambda)_T$ ones in some settings. However, on \RING-4, all entropy values collapse to 0.0 regardless of population size, clearly indicating mode collapse and confirming that some evolutionary dynamics are necessary to preserve diversity in structured distributions.

\begin{table}[h!]
\captionsetup{position=top, skip=2pt}.
\renewcommand{\arraystretch}{0.9}  % or 0.8 for more compact

    \centering
    \caption{Mode coverage median (iqr) for $(\mu+\lambda)_E$ and $(\mu+\lambda)_T$ on \BLOB-8 (ideal value 8) and \RING-4 (ideal value 4).} 
    \label{tab:modes-mu-plus-lambda}
    \begin{tabular}{
        c@{\hskip 4pt}
        c@{\hskip 6pt}c
        c@{\hskip 4pt}
        c@{\hskip 10pt}
        c@{\hskip 4pt}
        c
    }
        \toprule
        
        & & \multicolumn{2}{c}{\BLOB-8} &\phantom{---} & \multicolumn{2}{c}{\RING-4} \\
        \cmidrule{3-4} \cmidrule{6-7}  
        $\mu$ & $\lambda$
        & $(\mu+\lambda)_E$ & $(\mu+\lambda)_T$ &\phantom{---}
        & $(\mu+\lambda)_E$ & $(\mu+\lambda)_T$ \\
        \midrule
        3 & 1 & 5.5 (6.0) & 4.0 (2.25) &\phantom{---} & 2.0 (3.00) & 2.0 (1.00) \\
        3 & 2 & 5.5 (6.0) & 8.0 (6.00) &\phantom{---} & 2.0 (2.25) & 2.0 (3.00) \\
        3 & 3 & 8.0 (4.0) & 8.0 (0.00) &\phantom{---} & 2.0 (1.00) & 2.0 (3.00) \\
        5 & 1 & 6.0 (4.0) & 4.0 (1.25) &\phantom{---} & 2.0 (1.25) & 2.0 (1.00) \\
        5 & 3 & 8.0 (3.5) & 8.0 (5.25) &\phantom{---} & 2.0 (0.25) & 1.5 (3.00) \\
        5 & 5 & 5.0 (5.5) & 8.0 (1.00) &\phantom{---} & 2.0 (1.00) & 1.5 (3.00) \\
        7 & 1 & 5.0 (1.5) & 3.0 (1.50) &\phantom{---} & 2.0 (2.00) & 2.0 (0.50) \\
        7 & 4 & 7.0 (4.5) & 8.0 (6.00) &\phantom{---} & 2.0 (2.25) & 2.0 (1.00) \\
        7 & 7 & 8.0 (1.0) & 8.0 (5.00) &\phantom{---} & 2.0 (3.00) & 1.0 (1.25) \\
        \bottomrule
    \end{tabular}
\end{table}

\begin{table}[h!]
\captionsetup{position=top, skip=2pt}.
\renewcommand{\arraystretch}{0.9}  % or 0.8 for more compact

    \centering
    \caption{Mode coverage median (iqr) for $(\mu,\lambda)$ on \BLOB-8 (ideal value 8) and \RING-4 (ideal value 4).} 
    \label{tab:modes-mu-lambda}
    \begin{tabular}{ccccc}
        \toprule
        $\mu$ & $\lambda$ & \BLOB-8 & \RING-4 \\
        \midrule
        3 & 3  & 8.0 (0.0)   & 3.5 (2.0) \\
        3 & 5  & 8.0 (0.0)   & 4.0 (0.25) \\
        3 & 6  & 8.0 (0.0)   & 4.0 (1.00) \\
        5 & 5  & 8.0 (0.0)   & 4.0 (1.25) \\
        5 & 8  & 8.0 (0.0)   & 4.0 (2.25) \\
        5 & 10 & 8.0 (0.0)   & 4.0 (3.00) \\
        7 & 7  & 8.0 (0.0)   & 4.0 (0.25) \\
        7 & 11 & 8.0 (0.0)   & 4.0 (2.50) \\
        7 & 14 & 8.0 (0.0)   & 4.0 (1.00) \\
        \bottomrule
    \end{tabular}
\end{table}

Mode coverage analysis reinforces these findings. As shown in Table~\ref{tab:modes-mu-lambda}, $(\mu, \lambda)$ achieves perfect mode coverage (median 8.0 on \BLOB-8, 4.0 on \RING-4) in all evaluated configurations, with negligible or zero variance. This indicates not only robust diversity but also full mode representation across all runs. In contrast, the $(\mu{+}\lambda)$ variants, especially $(\mu{+}\lambda)_E$, show frequent mode under-coverage and high variability. For instance, on \RING-4 with $\mu=7$, $\lambda=7$, $(\mu{+}\lambda)_E$ achieves a median mode count of only 2.0, while $(\mu{+}\lambda)_T$ drops further to 1.0 (Table~\ref{tab:modes-mu-plus-lambda}).

MG-MD GAN reaches full mode coverage (8.0) on \BLOB-8 for all population sizes. However, it fails completely on \RING-4, with a consistent median of 1.0. This further emphasizes that although static diversity can be sufficient in unstructured datasets, more targeted evolutionary mechanisms are required to handle complex data distributions with multiple tightly clustered modes.

\begin{table*}[h!]
\captionsetup{position=top, skip=2pt}.
\centering
\caption{Median FID (iqr) for each experiment variant by $\mu$ and $\lambda$.}
\label{tab:fid_summary}
\begin{tabular}{lccccccc}
\toprule
$\mu$ & $(\mu+\lambda)_E$ $\lambda$=1 & $(\mu+\lambda)_E$ $\lambda$=$\mu$ & $(\mu+\lambda)_T$ $\lambda$=1 & $(\mu+\lambda)_T$ $\lambda$=$\mu$ & $(\mu,\lambda)$ $\lambda$=$\mu$ & $(\mu,\lambda)$ $\lambda$=2$\mu$ & MG-MD GAN\\
\midrule
3 & 55.463 (1.645) & 52.321 (2.588) & 53.488 (3.282) & 50.178 (1.715) & 47.156 (0.668) & 45.218 (0.825) & 48.357 (1.551) \\
5 & 52.760 (1.309) & 50.221 (1.888) & 52.030 (2.234) & 48.475 (1.657) & 46.145 (1.063) & 44.074 (1.034) & 46.622 (1.678) \\
\bottomrule
\end{tabular}

\end{table*}

\begin{table*}[t!]
\caption{Median TVD (iqr) for each experiment variant by $\mu$ and $\lambda$.}
\label{tab:tvd_summary}
\centering
\begin{tabular}{lccccccc}
\toprule
$\mu$ & $(\mu+\lambda)_E$ $\lambda$=1 & $(\mu+\lambda)_E$ $\lambda$=$\mu$ & $(\mu+\lambda)_T$ $\lambda$=1 & $(\mu+\lambda)_T$ $\lambda$=$\mu$ & $(\mu,\lambda)$ $\lambda$=$\mu$ & $(\mu,\lambda)$ $\lambda$=2$\mu$ & MG-MD GAN\\
\midrule
3 & 1.975 (0.383) & 1.822 (0.221) & 1.901 (0.450) & 1.689 (0.255) & 1.633 (0.516) &  1.518 (0.362) &  1.646 (0.263) \\
5 & 1.943 (0.224) & 1.743 (0.238) & 1.636 (0.386) & 1.590 (0.433) & 1.475 (0.320) & 1.467 (0.206)) & 1.603 (0.284) \\
\bottomrule
\end{tabular}
\end{table*}

\begin{figure}[!ht]
\centering
\begin{subfigure}[t]{0.24\linewidth}
   \centering
    \includegraphics[width=0.99\linewidth]{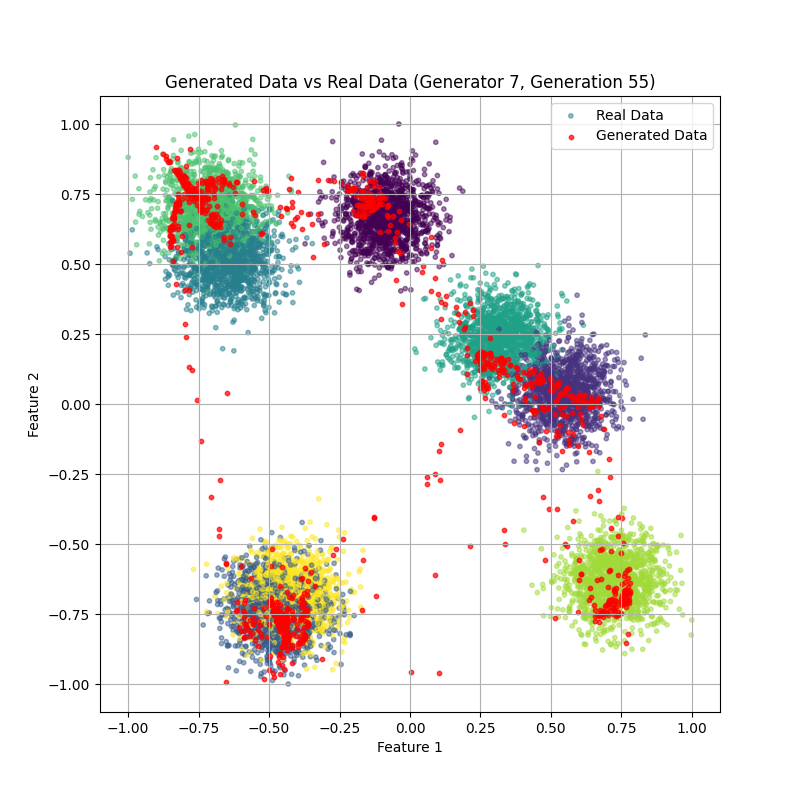}
    
    \caption{$(\mu+\lambda)_E$}
\end{subfigure}
\begin{subfigure}[t]{0.24\linewidth}
   \centering
    \includegraphics[width=0.99\linewidth]{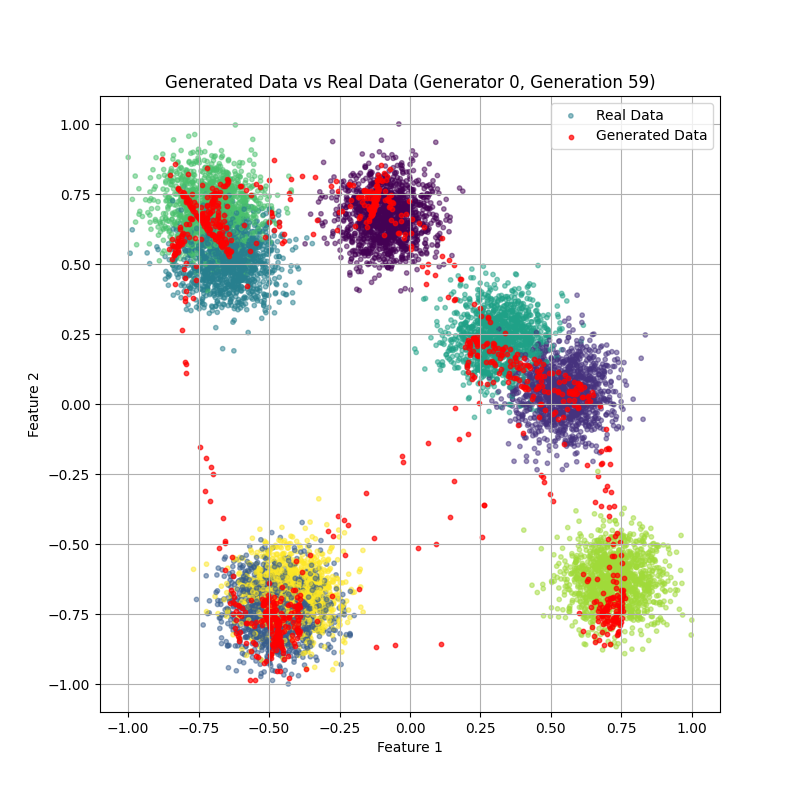}
    \caption{$(\mu+\lambda)_T$}
\end{subfigure}
\begin{subfigure}[t]{0.24\linewidth}
   \centering
    \includegraphics[width=0.99\linewidth]{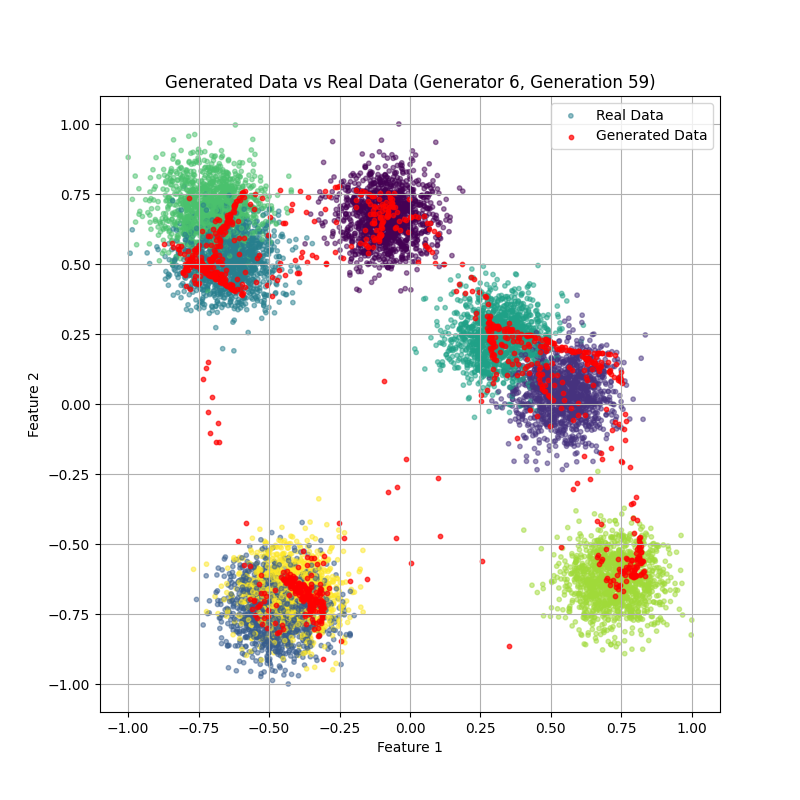}
    \caption{$(\mu,\lambda)$}
\end{subfigure}
\begin{subfigure}[t]{0.24\linewidth}
   \centering
    \includegraphics[width=0.99\linewidth]{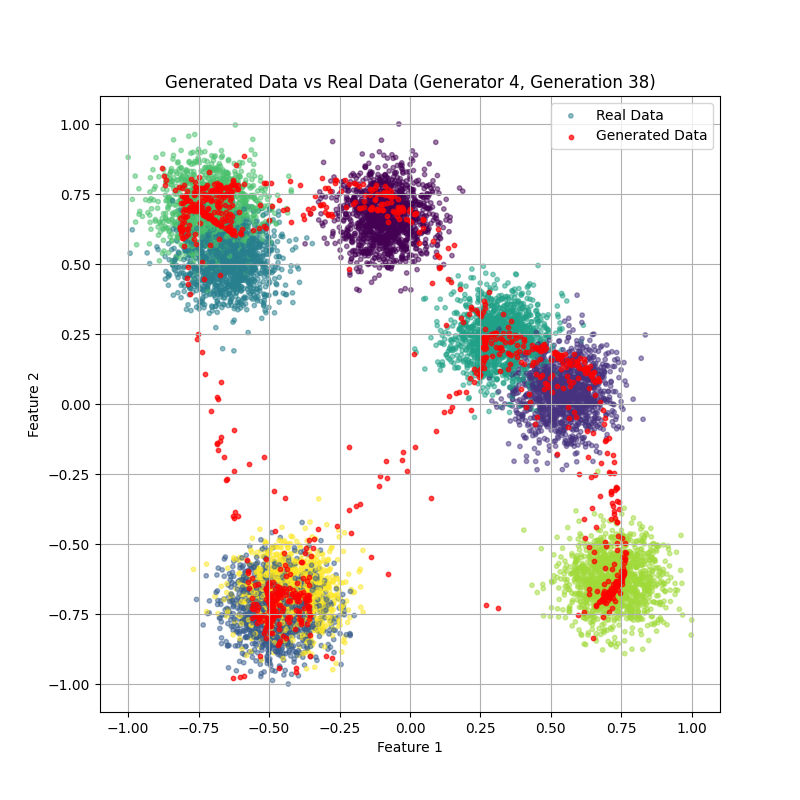}
    \caption{MG-MD GAN}
    \label{fig:ring-wd-population}
\end{subfigure}
        \vspace{-0.8em} % Reduce space between image and caption
\caption{\BLOB-8 samples produced when using $\mu$=3.}
\label{fig:blobs-8-samples}
\end{figure}

\begin{figure}[!ht]
\centering
\begin{subfigure}[t]{0.24\linewidth}
   \centering
    \includegraphics[width=0.99\linewidth]{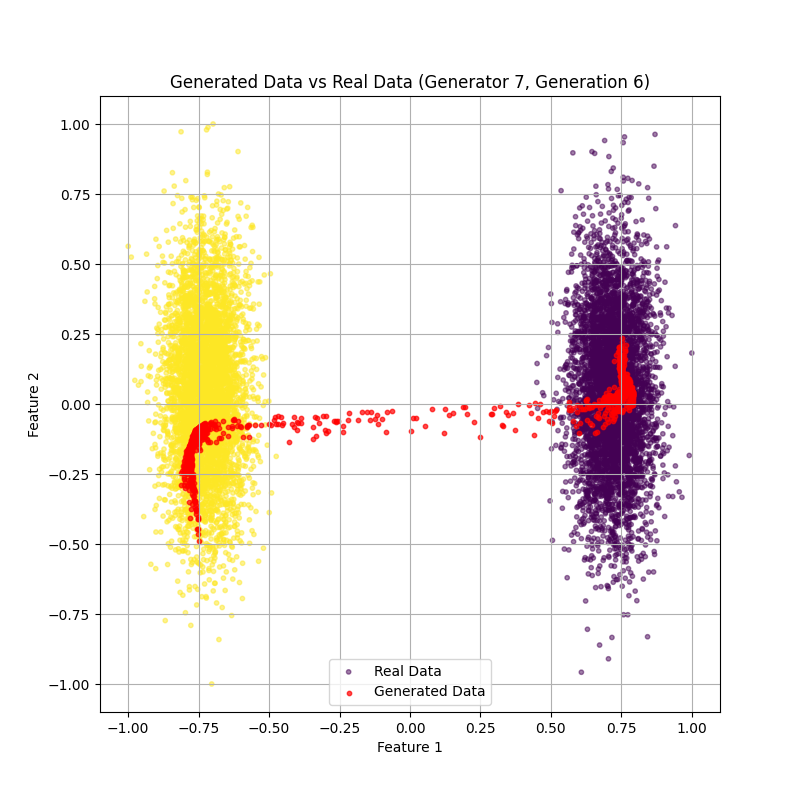}
    \caption{$(\mu+\lambda)_E$}
\end{subfigure}
\begin{subfigure}[t]{0.24\linewidth}
   \centering
    \includegraphics[width=0.99\linewidth]{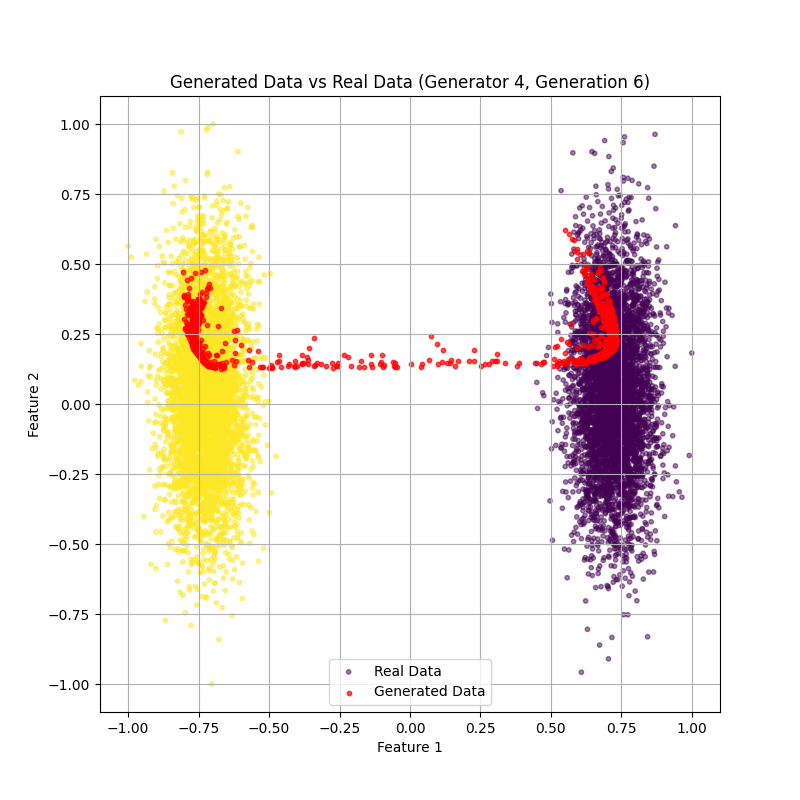}
    \caption{$(\mu+\lambda)_T$}
\end{subfigure}
\begin{subfigure}[t]{0.24\linewidth}
   \centering
    \includegraphics[width=0.99\linewidth]{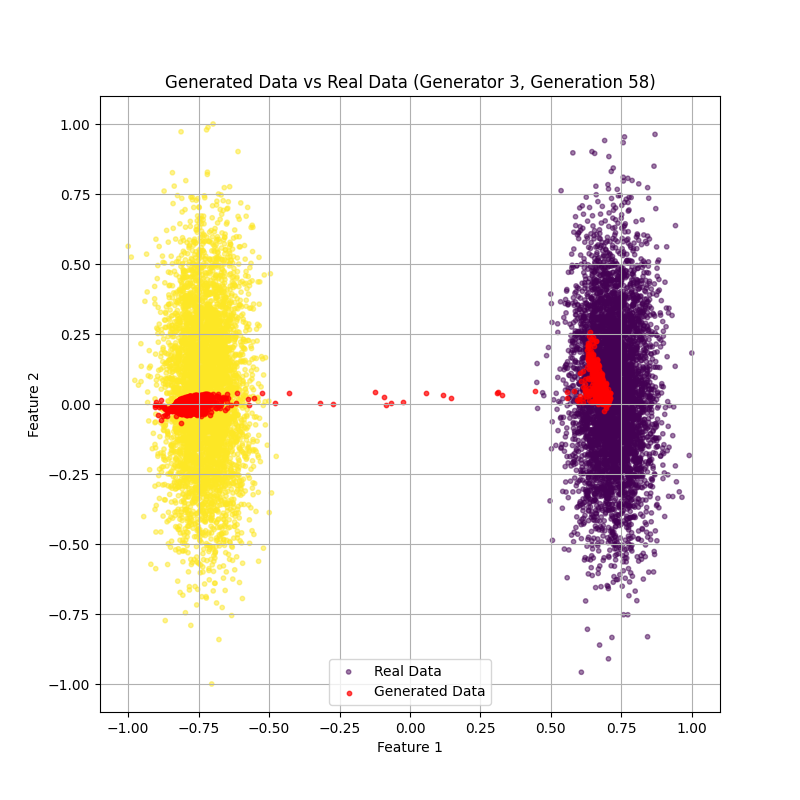}
    \caption{$(\mu,\lambda)$}
\end{subfigure}
\begin{subfigure}[t]{0.24\linewidth}
   \centering
    \includegraphics[width=0.99\linewidth]{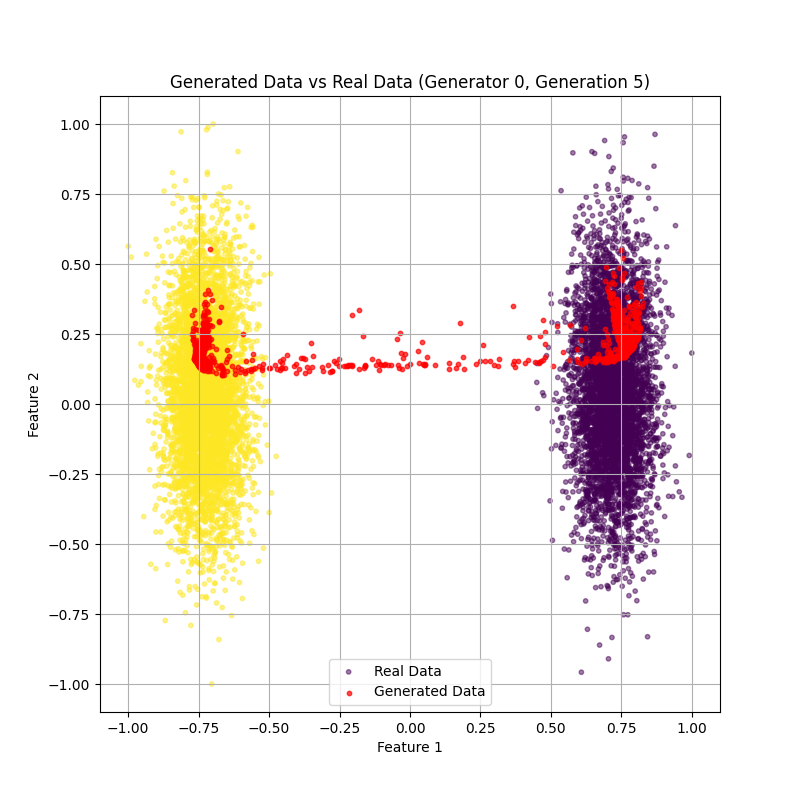}
    \caption{MG-MD GAN}
    \label{fig:ring-wd-population}
\end{subfigure}
        \vspace{-0.8em} % Reduce space between image and caption

\caption{\RING-2 samples produced when using $\mu$=3.}
\label{fig:gaussian-2-samples}
\end{figure}

\begin{figure}[!ht]
\centering
\begin{subfigure}[t]{0.24\linewidth}
   \centering
    \includegraphics[width=0.99\linewidth]{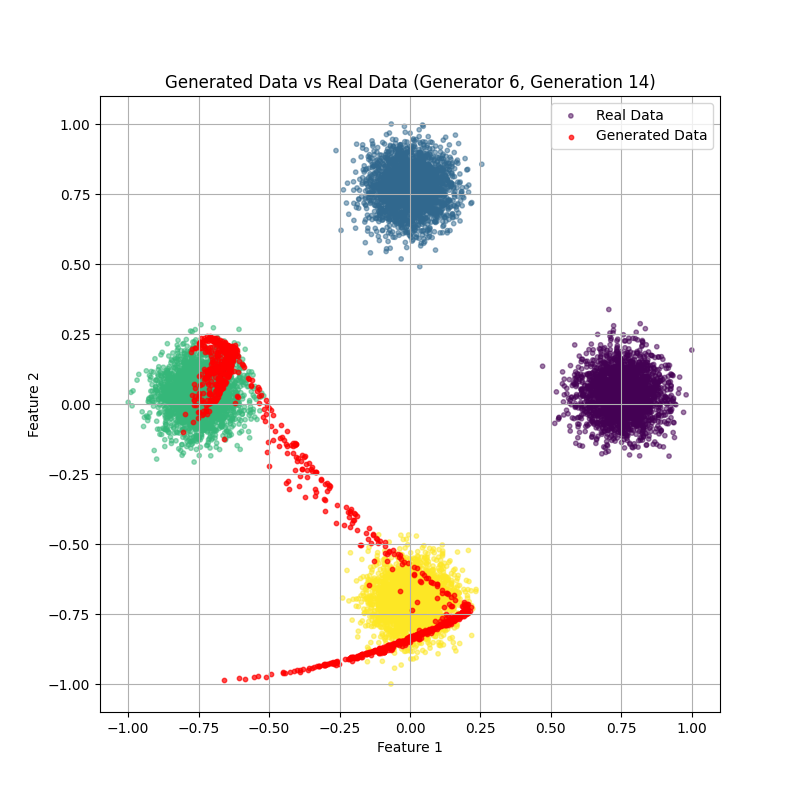}
    \caption{$(\mu+\lambda)_E$}
\end{subfigure}
\begin{subfigure}[t]{0.24\linewidth}
   \centering
    \includegraphics[width=0.99\linewidth]{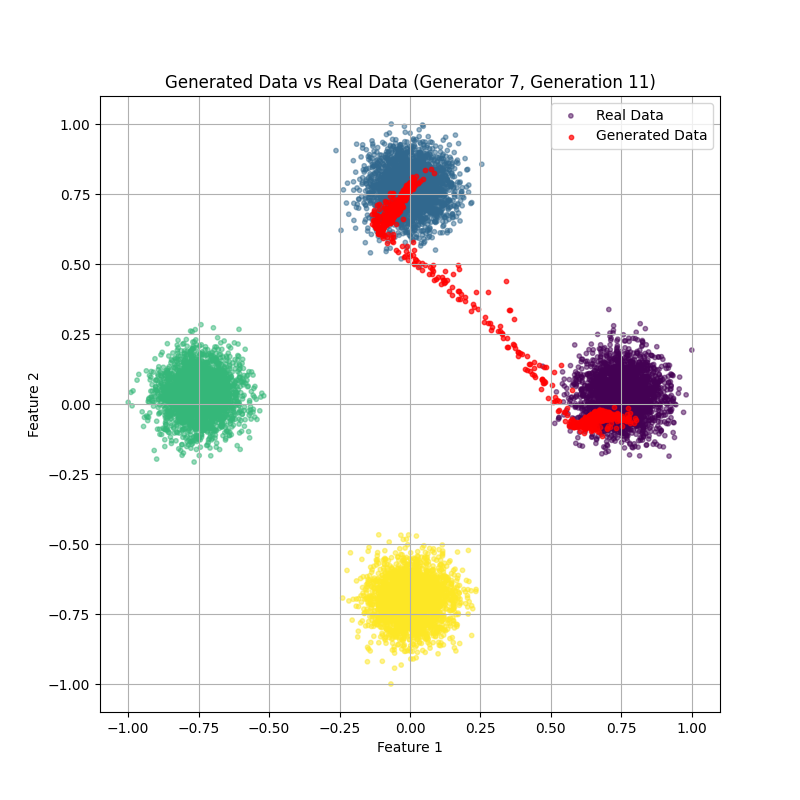}
    \caption{$(\mu+\lambda)_T$}
\end{subfigure}
\begin{subfigure}[t]{0.24\linewidth}
   \centering
    \includegraphics[width=0.99\linewidth]{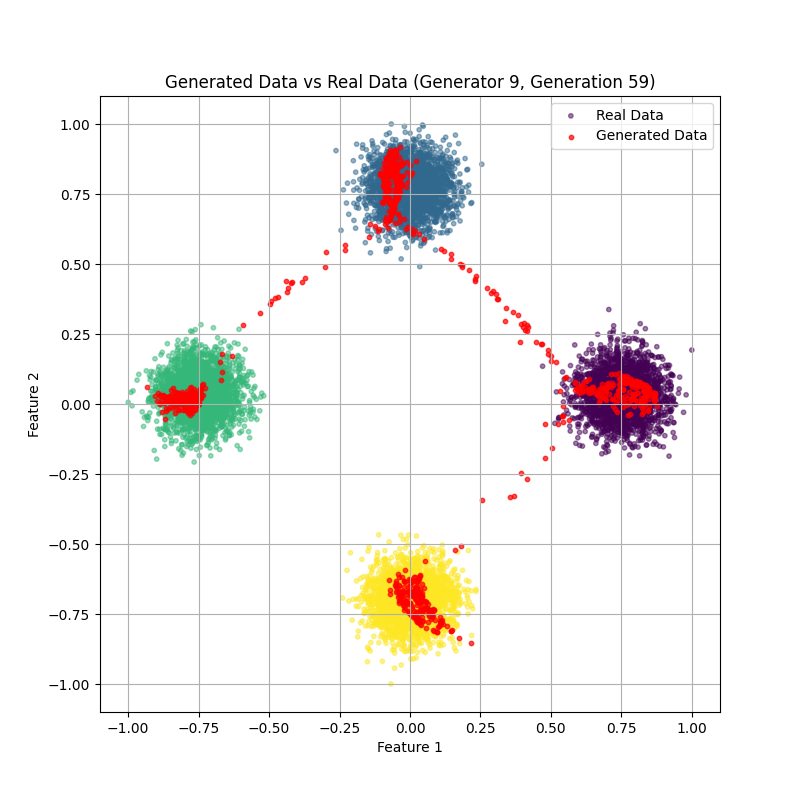}
    \caption{$(\mu,\lambda)$}
\end{subfigure}
\begin{subfigure}[t]{0.24\linewidth}
   \centering
    \includegraphics[width=0.99\linewidth]{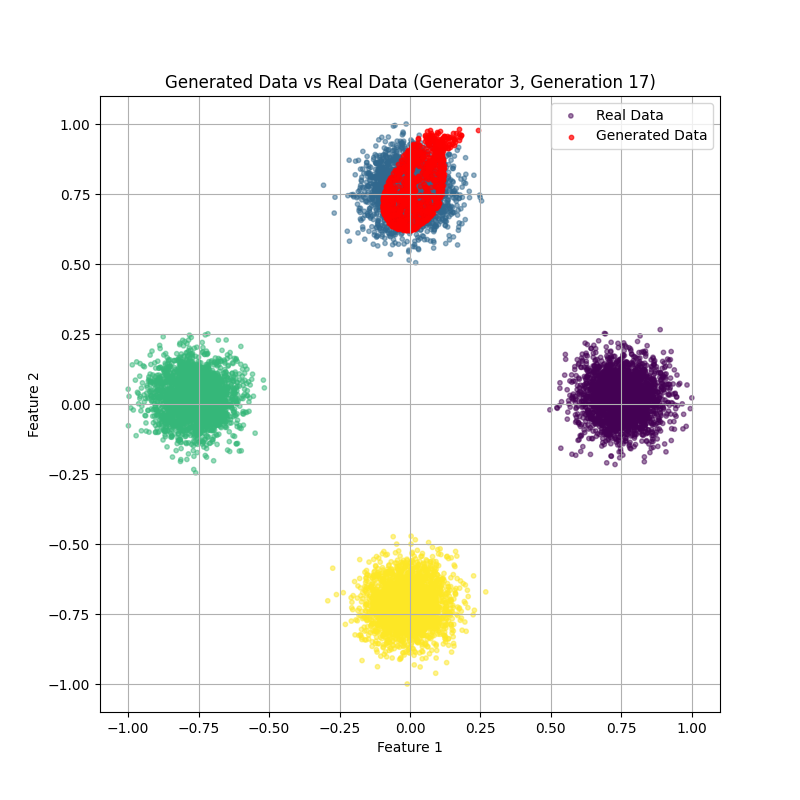}
    \caption{MG-MD GAN}
    \label{fig:ring-wd-population}
\end{subfigure}
        \vspace{-0.8em} % Reduce space between image and caption

\caption{\RING-4 samples produced when using $\mu$=3.}
\label{fig:gaussian-4-samples}
\end{figure}

\begin{figure}[!ht]
\centering
\begin{subfigure}[t]{0.24\linewidth}
   \centering
    \includegraphics[width=0.99\linewidth]{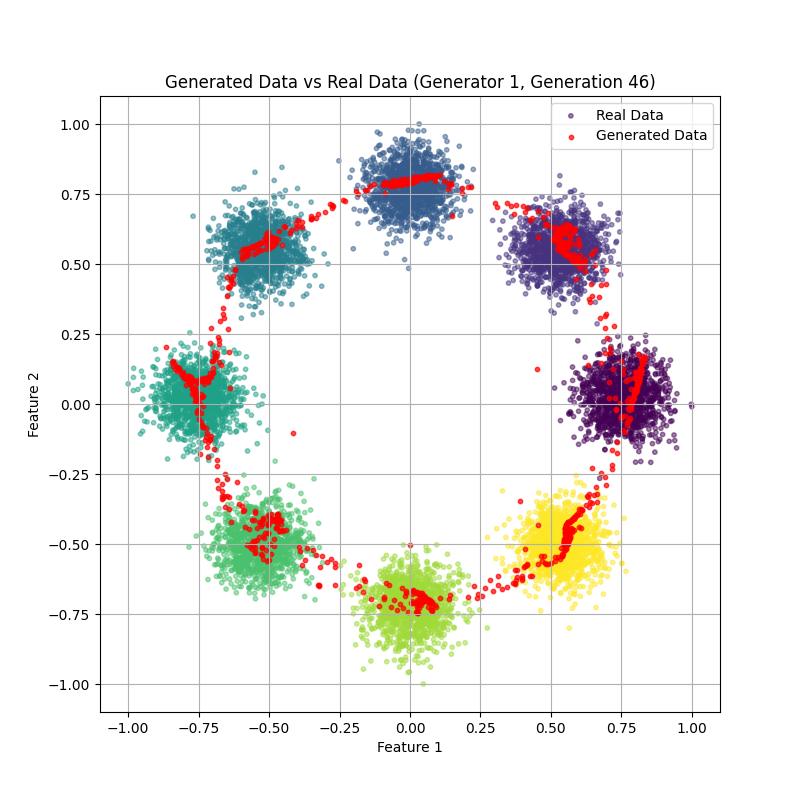}
    \caption{$(\mu+\lambda)_E$}
\end{subfigure}
\begin{subfigure}[t]{0.24\linewidth}
   \centering
    \includegraphics[width=0.99\linewidth]{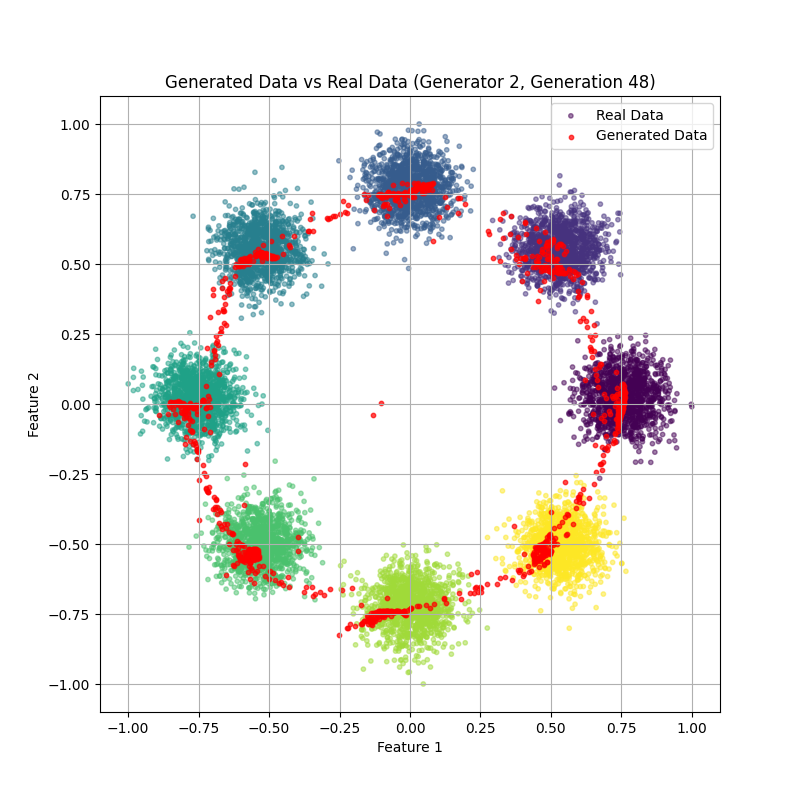}
    \caption{$(\mu+\lambda)_T$}
\end{subfigure}
\begin{subfigure}[t]{0.24\linewidth}
   \centering
    \includegraphics[width=0.99\linewidth]{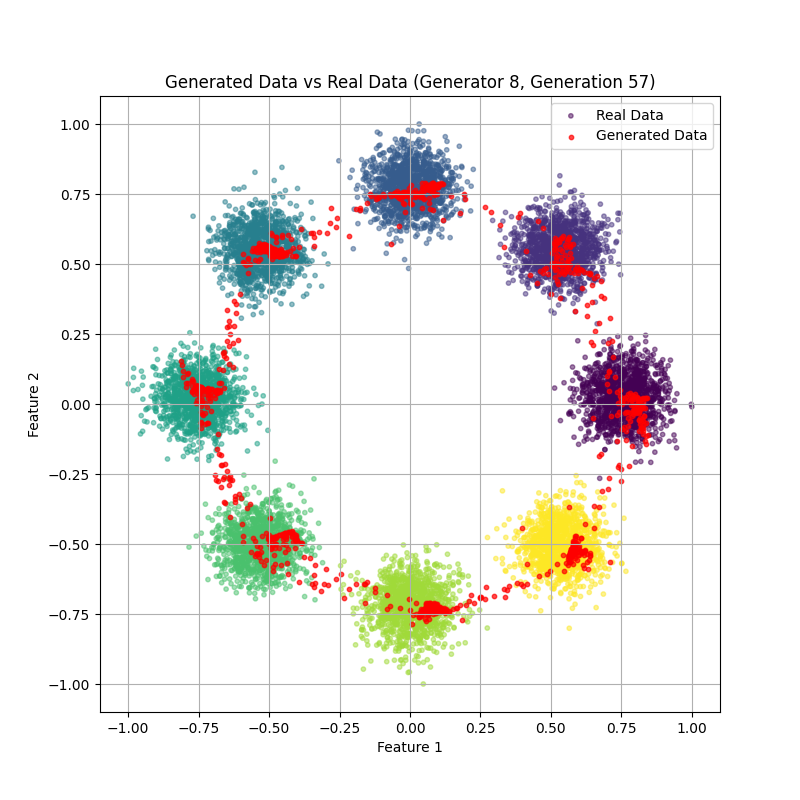}
    \caption{$(\mu,\lambda)$}
\end{subfigure}
\begin{subfigure}[t]{0.24\linewidth}
   \centering
    \includegraphics[width=0.99\linewidth]{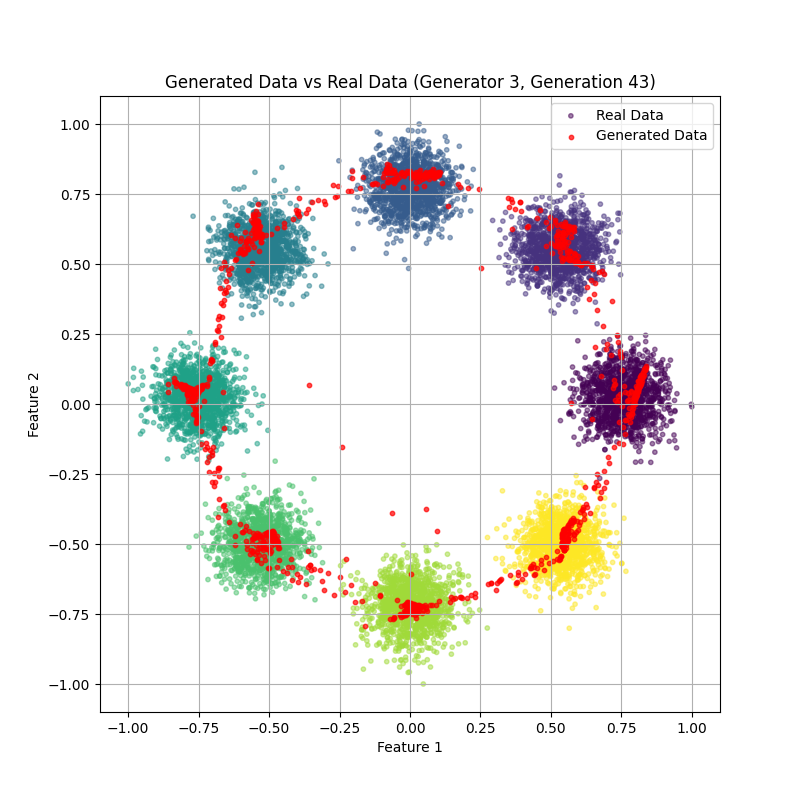}
    \caption{MG-MD GAN}
    \label{fig:ring-wd-population}
\end{subfigure}
        \vspace{-0.8em} % Reduce space between image and caption
\caption{\RING-8 samples produced when using $\mu$=3.}
\label{fig:gaussian-8-samples}
\end{figure}

Figures~\ref{fig:blobs-8-samples},~\ref{fig:gaussian-2-samples},~\ref{fig:gaussian-4-samples}, and~\ref{fig:gaussian-8-samples} present a representative set of the synthetic samples created by each one of the methods for \BLOB-8, \RING-2, \RING-4, and~\RING-8 datasets, respectively. In these figures, red dots represent the produced samples.

A general trend can be seen in the visual results in figures~\ref{fig:blobs-8-samples}-\ref{fig:gaussian-8-samples} in which $(\mu,\lambda)$ GAN produced samples closer to the real dataset with fewer samples out of the modes that define the datasets than the other approaches. 

Focusing on the most challenging dataset.,i.e., \RING-4, the differences between the samples produced by $(\mu,\lambda)$ method is remarkable, since it was capable to produce samples for all the four modes (see Figure~\ref{fig:gaussian-4-samples}). In contrast, $(\mu+\lambda)$ variations collapse in two modes and MG-MD GAN collapses in only one mode.

\sloppy
Taken together, these results demonstrate that while non-evolutionary methods can yield competitive quality and mode coverage on simpler datasets, coevolutionary approaches with full generational replacement (i.e., $(\mu, \lambda)$) provide the most consistent and reliable strategy for maintaining diversity and avoiding mode collapse across a wide range of configurations and dataset complexities.

\subsection{Results on MNIST}

To limit computational cost in MNIST, we focus on the most promising configurations identified in the previous: population sizes $\mu \in {3, 5}$ and offspring sizes $\lambda \in \{1, \mu\}$ for $(\mu{+}\lambda)$ and $\lambda \in \{\mu, 2\mu\}$ for $(\mu,\lambda)$. These settings balance exploration and exploitation while remaining computationally feasible.
The resulting FID and TVD scores are 
summarized numerically in table~\ref{tab:fid_summary} and~\ref{tab:tvd_summary}, which reports the median and interquartile range (iqr) for each configuration.

The results reveal several notable trends. First, the $(\mu,\lambda)$ method, which implements a coevolutionary GAN training strategy based on the $(\mu,\lambda)$ evolutionary algorithm without elitism, consistently achieves the lowest FID scores, indicating the best generative performance overall. Across both $\mu=3$ and $\mu=5$, it shows strong medians and very narrow IQRs, suggesting not only high-quality outputs but also robust and stable behavior across runs.

In contrast, the $(\mu+\lambda)_E$ method, which applies pure elitism, while occasionally producing competitive results, generally yields higher FID scores, particularly for smaller values of $\lambda$. Although elitism is often considered beneficial for preserving top individuals in evolutionary computation, these results suggest that in the context of coevolutionary GAN training, it may introduce premature convergence or limit exploration, especially in the early training stages. The $(\mu+\lambda)_T$ variant, using tournament selection, performs similarly to the elitist variant but with greater variability.

The MG-MD GAN approach, which trains populations of generators and discriminators without using evolutionary algorithms, performs reasonably well in some configurations but does not match the performance of the $(\mu,\lambda)$ method. This further supports the benefit of explicitly applying evolutionary search dynamics in coevolutionary GAN training.

A pairwise Wilcoxon signed-rank test with Bonferroni correction was conducted to assess statistical significance. The results confirm that $(\mu,\lambda)$ is the most competitive method, with significantly better FID scores than most other configurations. In particular, the variant with $\mu=5$, $\lambda=10$ achieved the lowest overall median. Conversely, the $(\mu+\lambda)_E$ approach showed statistically significant worse performance in several pairwise comparisons, especially when $\lambda$ was small.

In terms of diversity, the TVD results in Table~\ref{tab:tvd_summary} and the Wilcoxon signed-rank test with Bonferroni correction show that the $(\mu, \lambda)$ outperforms the other methods. These variants got the lowest TVD scores overall, indicating that the distribution of generated digits closely matches the real data label distribution. Specifically, the $(\mu{=}5,\lambda{=}10)$ configuration achieves the best diversity result with a median TVD of 1.467, followed closely by $(\mu{=}5,\lambda{=}5)$, $(\mu{=}3,\lambda{=}6)$, and $(\mu{=}3,\lambda{=}3)$, confirming a clear advantage for full generational replacement strategies. 

The $(\mu{+}\lambda)$ variations generally obtained higher TVD scores and wider interquartile ranges, reflecting less consistent alignment between the generated and real data distributions. Using tournament selection outperforms the elitist approach. This improvement can be attributed to the greater exploration enabled by tournament selection, which fosters population diversity and reduces the risk of premature convergence. Although some $(\mu{+}\lambda)_T$ configurations approach the performance of $(\mu,\lambda)$ in specific settings, e.g., $(\mu{=}5,\lambda{=}\mu)$,, their overall results remain less robust. The MG-MD GAN baseline also performs competitively in some cases but falls short of the best $(\mu,\lambda)$ variants, reinforcing the importance of evolutionary dynamics for maintaining class-level diversity in image generation tasks.

Besides, all evaluated methods achieved the ideal mode coverage of 10, indicating that each digit class was represented in the generated samples. However, the TVD results highlight that merely covering all classes is not sufficient, the generated distributions must also approximate the real data distribution closely. As it has been discussed in the \BLOB and \RING results, these findings underline the critical role of population diversity and controlled evolutionary pressure in avoiding mode imbalance and ensuring stable, high-quality generation across all output classes.

\begin{figure}[!ht]
\centering

\begin{subfigure}[t]{0.35\linewidth}
   \centering
    \includegraphics[width=0.99\linewidth]{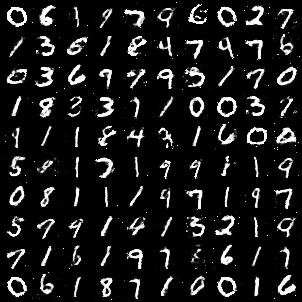}
    \caption{$(\mu+\lambda)_E$}
\end{subfigure}
\hspace{0.5cm}
\begin{subfigure}[t]{0.35\linewidth}
   \centering
    \includegraphics[width=0.99\linewidth]{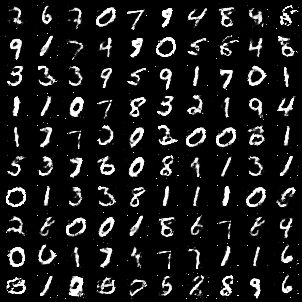}
    \caption{$(\mu+\lambda)_T$}
\end{subfigure}
\vspace{0.1cm}

\begin{subfigure}[t]{0.35\linewidth}
   \centering
    \includegraphics[width=0.99\linewidth]{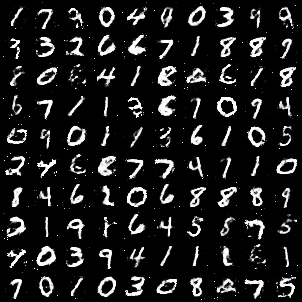}
    \caption{$(\mu,\lambda)$}
\end{subfigure}
\hspace{0.5cm}
\begin{subfigure}[t]{0.35\linewidth}
   \centering
    \includegraphics[width=0.99\linewidth]{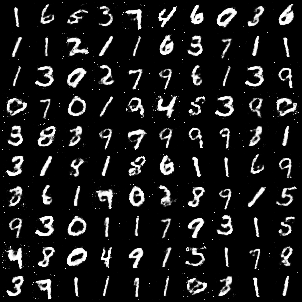}
    \caption{MG-MD GAN}
    \label{fig:ring-wd-population}
\end{subfigure}
\caption{MNIST samples produced when using $\mu$=3.}
\label{fig:mnist-samples}
\end{figure}

Figure~\ref{fig:mnist-samples} shows some samples produced by the analyzed approaches when using population size 3. It can be observed that in general all the approaches produced high quality samples. However, $(\mu+\lambda)_E$ samples set has lower diversity, i.e., high frequency of one digit, than the set produced by $(\mu,\lambda)$.

\subsection{General Discussion}

This section discussed the main findings in the context of the research questions introduced in Section~\ref{sec:introduction}.

\textbf{RQ1:} How do different coevolutionary strategies—particularly selection and replacement mechanisms—impact the quality and diversity of GAN-generated samples? 
The experiments show that the choice of selection and replacement strategy has a major effect on both quality and diversity. Full generational replacement strategies, i.e., $( \mu, \lambda )$, consistently outperform elitist and tournament-based $( \mu{+}\lambda )$. Elitism promotes exploitation but often leads to premature convergence and mode collapse. In contrast, tournament selection combined with generational replacement fosters diversity and leads to better and more stable outcomes.

\textbf{RQ2:} To what extent does the balance between exploration and exploitation, as controlled by population and offspring sizes, affect the performance and stability of GAN training?  
Increasing offspring size generally improves performance by enhancing exploratory capacity, particularly in tournament-based methods. However, large populations alone are not sufficient. The best results are obtained when larger offspring sizes are combined with full replacement, allowing broader exploration while avoiding dominance by early high performers. These effects are especially evident in structured datasets like \RING-4 and MNIST.

\textbf{RQ3:} Can a simple population based multi-generator multi-discriminator (MG-MD) GAN without selection or replacement be competitive with coevolutionary approaches?
MG-MD GAN performs competitively on simpler datasets, e.g., \BLOB-8, and achieves full mode coverage. However, it fails to generalize to more structured or complex distributions, such as \RING-4 and MNIST, where it suffers from mode collapse and reduced diversity. These results suggest that while static diversity can help in simple domains, evolutionary dynamics are necessary to achieve robustness and generalization in more challenging scenarios.

%%%%%%%%%%%%%%%%%%%%%%%%%%%%%%%%%%%%%%%%
%%%% -----------------------------------
%%%% --- CONCLUSIONS
%%%% -----------------------------------
%%%%%%%%%%%%%%%%%%%%%%%%%%%%%%%%%%%%%%%%

\section{Conclusions and Future Work}
\label{sec:conclusions}

This paper presents an empirical analysis of coevolutionary strategies for GAN training. We studied the impact of selection and replacement mechanisms across multiple configurations, comparing full generational replacement, i.e., $(\mu,\lambda)$, and elitist and tournament-based $(\mu{+}\lambda)$) approaches, along with a non-evolutionary MG-MD baseline. The evaluation was conducted on both synthetic 2D datasets and the MNIST image dataset.

The results consistently show that $(\mu,\lambda)$ based variations outperform other variants in terms of both sample quality and diversity. These configurations encourage exploration by discarding parents and replacing them entirely with offspring, which prevents premature convergence and promotes robustness. Tournament-based selection improves over elitism, but only when coupled with sufficient exploration via offspring replacement. While the MG-MD baseline performs well on simple datasets, it fails to generalize to more structured or high-dimensional data, highlighting the importance of applying evolutionary pressure.

Future work will explore more complex datasets (e.g., CIFAR-10, SVHN), other coevolutionary mechanisms such as spatially structured populations or quality-diversity methods, and multi-objective formulations that explicitly balance quality and diversity. Additionally, we aim to extend this framework to assess coevolutionary dynamics in conditional and multi-modal generative models.

\bibliographystyle{ACM-Reference-Format}
\bibliography{sample-base}
\end{document}